\documentclass{article} 

\usepackage[T1]{fontenc}
\usepackage{hyperref}
\usepackage{amsmath}
\usepackage{amssymb}
\usepackage{mathtools}
\usepackage{amsthm}
\usepackage{algorithm}
\usepackage{algorithmic}
\usepackage{booktabs}
\usepackage{pifont}
\usepackage{graphicx,wrapfig,lipsum}
\usepackage{microtype}
\usepackage{subfigure}
\usepackage[capitalize,noabbrev]{cleveref}
\usepackage[final]{neurips_2022}
\usepackage{times}
\usepackage{xcolor}
\title{Open-Ended Reinforcement Learning with Neural Reward Functions}

\author{Robert Meier \thanks{equal contribution} \\
Department of Computer Science\\
ETH Zürich\\
Zürich, Switzerland \\
\texttt{romeier@inf.ethz.ch} \\
\And
Asier Mujika $^*$ \\
Department of Computer Science\\
ETH Zürich\\
Zürich, Switzerland \\
\texttt{asierm@inf.ethz.ch} \\
}

\newcommand{\ant}{\textsc{Ant} }
\newcommand{\hum}{\textsc{Humanoid} }
\newcommand{\cheetah}{\textsc{Half-Cheetah} }
\newcommand{\diayn}{\textsc{DIAYN} }

\newcommand{\gcrl}{\textsc{GCRL} }

\let\cite\citep
\begin{document}

\maketitle

\begin{abstract}
Inspired by the success of unsupervised learning in Computer Vision and Natural Language Processing, the reinforcement learning community has recently started  to focus more on unsupervised discovery of skills.
Most current approaches, like DIAYN or DADS, optimize some form of mutual information objective.
We propose a different approach that uses reward functions encoded by neural networks.
These are trained iteratively to reward more complex behavior.
In high-dimensional robotic environments our approach learns a wide range of interesting skills including front-flips for \cheetah  and one-legged running for \textsc{Humanoid}.
It is the first skill discovery algorithm that can learn such skills without relying on any form of feature engineering.
In the pixel-based Montezuma's Revenge environment our method also works with minimal changes and it learns complex skills that involve interacting with items and visiting diverse locations.
The implementation of our approach can be found  \href{https://github.com/amujika/Open-Ended-Reinforcement-Learning-with-Neural-Reward-Functions}{\color{blue}{here}}.

\end{abstract}

\section{Introduction}

Deep reinforcement learning (RL) has proven to be very successful in many challenging tasks \cite{mnih2015human, silver2017mastering, berner2019dota}. These were considered intractable just a few years ago.
However, current methods require enormous amounts of compute to achieve great performance in individual tasks.
Most of the time, previous models can not be utilised when new tasks are considered, even if the environment does not change.
This was also the case in Computer Vision and Natural Language Processing.
However, recently unsupervised learning has been shown to be very effective in both fields.
By using task agnostic pre-training schemes \cite{tenney2019bert, brown2020language, caron2021emerging}, unsupervised models can solve most tasks with minimal or even no fine-tuning.
Unsupervised reinforcement learning aims to bring similar successes to the reinforcement learning community.

Agents which learn a wide range of unsupervised skills may be able to leverage those to solve new tasks faster and with minimal fine-tuning.

Here, we propose a new method for open-ended, unsupervised skill discovery. 

We devise an iterative process which creates pairs of neural reward functions and policies. The policy optimizes the corresponding reward function. 
Each of them corresponds to a skill that the agent has learned.
The neural reward function is a neural network that maps the current observation to a scalar reward. 
In each iteration, the neural reward function is modified to differ from the previous one. This results in increasing the complexity of the encoded task. 
Then, a new skill is trained to optimize this reward function. 
We devise several techniques to transfer the knowledge from previously learned skills.
These mechanisms enable learning the most complex reward functions that our method creates.
In fact, we show that some of the functions are impossible to learn from scratch.

We empirically test our framework in a diverse set of environments. 
First, we apply it to a simple 2d navigation task.
This lets us perform experiments quickly and gain a solid empirical understanding of the different components of our method.
Then, we apply it to three robotic environments where we have continuous high-dimensional observations and actions. 
In contrast to previous methods, our approach can deal with high-dimensional input directly without feature engineering. 
Finally, we apply it on the challenging Montezuma's Revenge Atari game which has visually rich pixel based observations and discrete actions. 
Despite their differences, our method manages to learn useful and interesting skills in all of them.
This shows that neural reward functions are equipped to encode meaningful tasks in diverse environments. 
We highlight the main contributions of this paper: 

\begin{itemize}
    \item Propose the first skill-discovery algorithm to work in high-dimensional environments without prior expert knowledge.
    \item Solve a 2d maze that cannot be solved by using random exploration.
    \item Acquire complex skills including performing front- and back-flip in the \textsc{Half-Cheetah}, running in all directions in \ant and \textsc{Humanoid}, standing and jumping on one leg in \textsc{Humanoid}.
    \item Learn to run as fast in \hum  as a supervised agent trained for tens of millions of steps.
    \item Achieve a higher Particle-Based Mutual Information Metric \cite{gu2021braxlines} than approaches that explicitly optimize this metric.
    \item Collect the first key and reach several rooms in Montezuma's Revenge.
\end{itemize}

\section{Related Work}

Our work fits best into the unsupervised skill discovery literature \cite{mohamed2015variational,gregor2016variational,florensa2017stochastic,achiam2018variational,eysenbach2018diversity,sharma2019dynamics,choi2021variational}. 
Compared to DIAYN \cite{eysenbach2018diversity} and similar approaches, we do not fix the number of skills to be learned at the beginning of training. 
This allows us to learn new skills in an open-ended fashion.
On top of that, maximizing mutual information can lead to degenerate behaviors in high-dimensional environments.
By manipulating a small subset of the dimensions, a lot of information can be encoded, without exploring the rest of the state space.
Due to this issue, previous methods only consider a hand-picked subset of the dimensions to perform well.

Open-ended learning \cite{srivastava2012continually,wang2019paired,wang2020enhanced,campero2020learning,dennis2020emergent,ecoffet2021first,team2021open} is closely related to unsupervised skill discovery.
However, most approaches require either a parameterizable environment \cite{wang2019paired,wang2020enhanced}, some fixed encoding of tasks \cite{team2021open} or self-competition \cite{silver2017self,baker2019emergent}.
This limits the applicability to environments that are engineered with these restrictions in mind.
In contrast, neural networks are universal function approximators and thus, our approach can encode any possible task in any possible environment, as long as the input is chosen appropriately. 
Similar to our method, the Go-Explore algorithm \cite{ecoffet2021first} also explores around the frontier of the known. It uses a handcrafted feature map to group similar states. Instead, we use neural reward functions to group them as skills. This does not use any expert knowledge.

Another related line of research to our approach is intrinsic motivation
\cite{stadie2015incentivizing,bellemare2016unifying,pathak2017curiosity,burda2018exploration,burda2018large,pathak2019self,raileanu2020ride}.
These approaches have managed great success in hard-exploration Atari games.
However, these approaches do not learn multiple skills by default. 
Additionally, their goal is to reward all novel states, which leads to general policies that visit many states.
In contrast, skill discovery algorithms try to reward narrow regions of the state space to achieve controllability of meaningful dimensions.

Encoding reward functions as neural networks has also been considered in the literature. Compared to our work, they are trained with supervised signals \cite{abbeel2004apprenticeship,fu2017learning,singh2019end,li2021mural}. Other approaches train auxiliary rewards with meta-learning \cite{zheng2018learning,du2019liir,veeriah2019discovery} to enhance the learning of the original reward function. 

Another approach to train multiple behaviors is goal-conditioned learning \cite{kaelbling1993learning,schaul2015universal,andrychowicz2017hindsight,rauber2017hindsight,nair2018visual,veeriah2018many,warde2018unsupervised,pong2019skew,choi2021variational}. In automated curriculum learning \cite{bengio2009curriculum,florensa2017reverse,forestier2017intrinsically,graves2017automated,sukhbaatar2017intrinsic,florensa2018automatic,matiisen2019teacher,narvekar2020curriculum,portelas2020teacher,portelas2020automatic,zhang2020automatic}, a sequence of goals is created such that each of them is not too hard nor too easy for the current agent.
These approaches mostly rely on low-dimensional goal embeddings.
When dealing with high-dimensional observations, they must use dimensionality reduction techniques.
These techniques can introduce instabilities or destroy relevant information from the input.
Our approach, on the other hand, can deal with high-dimensional inputs directly.
On top of that, goals encode a narrow region of the state space, while each of our reward functions can be rewarding in a large region.
This speeds-up the exploration in `easy' regions of the state space.

\section{Method}
\label{sec:methods}
\begin{wrapfigure}{r}{0.5\columnwidth}
\vspace{-1.5cm}
\includegraphics[width=0.5\columnwidth]{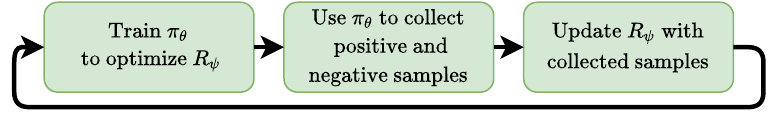}
\caption{Main steps of our algorithms open-ended training loop.}\label{fig:procedure}
\vspace{-0.1cm}
\end{wrapfigure}

We introduce a method that performs open-ended, unsupervised skill discovery. It iteratively creates pairs of neural reward functions $R_{\psi}$ and policies $\pi_{\theta}$ trained to maximize the corresponding $R_{\psi}$. 
Our proposed method alternates between increasing the complexity of the reward function $R_{\psi}$ and leveraging the previously learned skills to learn a policy $\pi_{\theta}$ that can solve the new $R_{\psi}$.
This yields a general learning procedure that learns complex skills in a diverse set of environments. See Figure~\ref{fig:procedure} for a high level overview.

\subsection{Increasing the complexity of $R_{\psi}$}
\label{sec:learning_R}

The reward function in a fully observable Markov Decision Process (MDP) is by definition a function of the current observation, the next one and the action that was performed.
However, in many cases this can be reduced to a function of just the current observation.
Because of this, we opt for these simpler reward functions as the basis of our $R_{\psi}$.
In our method, the reward function $R_{\psi}$ is a neural network which takes observations $o_t$ as input and outputs a single scalar value, the instantaneous reward $r^{\psi}_t$.

Assume that we have a policy $\pi_\theta$ that can reach states in the MDP which are rewarding under $R_{\psi}$. To increase the complexity of the reward function, we do the following:

\begin{itemize}
    \item \textbf{Decrease the reward} of states which are visited by $\pi_\theta$, as we already have a policy that reaches these states. We create a data set $O_{neg}$ of such states.  We refer to these states as \textit{negative samples}.
    \item \textbf{Increase the reward} of states that can almost be reached by the current policy. This allows us to leverage $\pi_\theta$ to learn the new reward function. We create a data set $O_{pos}$ of such states. We refer to these states as \textit{positive samples}.
\end{itemize}

To generate the negative samples, we run $\pi_\theta$ for a given number of steps (ideally until it reaches rewarding states) and store the visited states. 
To then generate positive samples, we change to performing random actions\footnote{In MDPs with discrete actions we sample actions u.a.r. and in the continuous case, we keep the mean of $\pi_\theta$ and increase the standard deviation.} for a fixed number of steps. 
 To ensure that the new reward function is different from all previous ones, we also keep track of all the negative samples that we have collected for all skills in a data set $O_{neg\_all}$.

Finally, we set target values $a$ and $-a$ for the positive and negative samples respectively, and train the reward network using standard supervised learning on the following loss:
\begin{align*}
   \mathcal{L}_\psi &= \sum_{o \in O_{neg}}  \frac{(R_\psi (o) + a)^2}{|O_{neg}|} +   \sum_{o \in O_{neg\_all}} \frac{(R_\psi (o) + a)^2 }{|O_{neg\_all}|}
  + \sum_{o \in O_{pos}}\frac{(R_\psi (o) - a)^2}{|O_{pos}|}
\end{align*}
This loss ensures that positive samples that have never been seen before have positive reward in the next $R_{\psi}$, while all other samples that have been seen before decrease their reward.
In the reinforcement learning phase we clip rewards to the $[0, a]$ range.
This ensures that the agent seeks positive samples, rather than less negative ones.

\subsection{Forward transfer for $\pi_\theta$}
\label{sec:method_transfer}

Given the procedure presented in Section \ref{sec:learning_R}, we create increasingly complex reward functions. 
While this is great for open-ended learning, it eventually leads to skills that are too complex and cannot be learned from scratch.
In order to learn these skills, we must leverage previous knowledge about the environment.
In this section we present several forward transfer mechanisms that are necessary for the most complex skills.

In principle, our method can be combined with any standard reinforcement learning (RL) technique. But here, we focus on actor-critic methods like Advantage Actor-Critic \cite{mnih2016asynchronous} or Proximal Policy Optimization \cite{schulman2017proximal} for learning $\pi_\theta$ for several reasons. 
These methods have a value network that is separate from the policy which allows us to use different transfer mechanisms for the value and policy network. 
Also, these approaches work for both continuous and discrete action spaces. This allows us to use the same technique for the robotic environments and for the 2d navigation tasks.
Finally the learned policies are stochastic which increases the diversity of negative samples and speeds up the skill discovery process. See Section \ref{sec:deep-exploration}.

We present our three forward transfer mechanisms below. All of them exploit the similarity between consecutive reward functions to ensure that even very complex reward functions can be solved by the RL agent in a reasonable number of environment interactions.

\begin{itemize}
    \item \textbf{Value Reuse:} Initialize the value network to the final value network of the previous agent. While two consecutive reward functions are different, both still reward close-by regions of the state space. Thus, by keeping the previous value function, the policy network will be nudged towards that region of the state space from the very first gradient updates.
    \item \textbf{Policy Feature Reuse:} Initialize the policy network to the final policy of the previous agent but setting the weights of the final layer to $0$. This keeps the previously learned features, but outputs a uniform policy over all actions (or mean $0$ and a fixed standard deviation in the continuous case) which allows for proper learning and exploration\footnote{Policies at the end of learning can be very deterministic which slows down or completely stops learning of new reward functions.}.
    \item \textbf{Guiding Policy:} Act with the previous policy for a random number of steps at the beginning of each episode. This heavily simplifies the exploration problem. 
    The agent will start exploring from states that are much closer to the rewards defined by $R_\psi$.
    This is because the previous policy could already solve the previous reward function.
    In contrast to the other two mechanisms, this one does not rely on initialization. Thus, this is the most effective in sparse reward functions that need many parameter updates to be learned.
\end{itemize}
    
In Section \ref{sec:navigation_forward} we individually evaluate these three techniques and show that their combination is necessary in complex environments.

Putting everything together we get an algorithm which learns reward functions that encode increasingly complex behaviors and learns RL agents that solve those reward functions.
Figure \ref{fig:procedure} illustrates the main steps of our training loop and see Algorithm \ref{alg:method} in the appendix for more detail.

\section{Experiments}
\label{sec:experiments}

We now proceed to experimentally test our method.
First, in Section \ref{sec:navigation}, we thoroughly test all different components of our model in a 2d navigation task. This task allows us to verify the function of each component and also to explicitly visualize what each reward function is encoding.

Then, in Section \ref{sec:robotics}, we move to BRAX robotic environments \cite{freeman2021brax}. These have a lot of flexibility and thus allow the agent to learn very complex tasks.
In these tasks, we evaluate the complexity of our skills by measuring their zero-shot transfer ability to the environment rewards. In the \hum environment, our unsupervised skills outperform supervised agents trained for tens of millions of time steps.
We also compute the one dimensional particle-based mutual information metric that has been proposed in the literature before \cite{gu2021braxlines} and show that our method outperforms previous approaches, even when other approaches only consider handcrafted feature dimensions in their objective. 

Finally, in Section \ref{sec:montezuma}, we apply our method to Montezuma's Revenge. We show that the learned reward functions keep getting increasingly complex and we are mostly limited by the amount of compute that it takes to learn each new reward function.

\subsection{2D Navigation Task}
\label{sec:navigation}

\begin{wrapfigure}{r}{0.4\columnwidth}
\vspace{-1.8cm}
\centerline{\includegraphics[width=0.35\columnwidth]{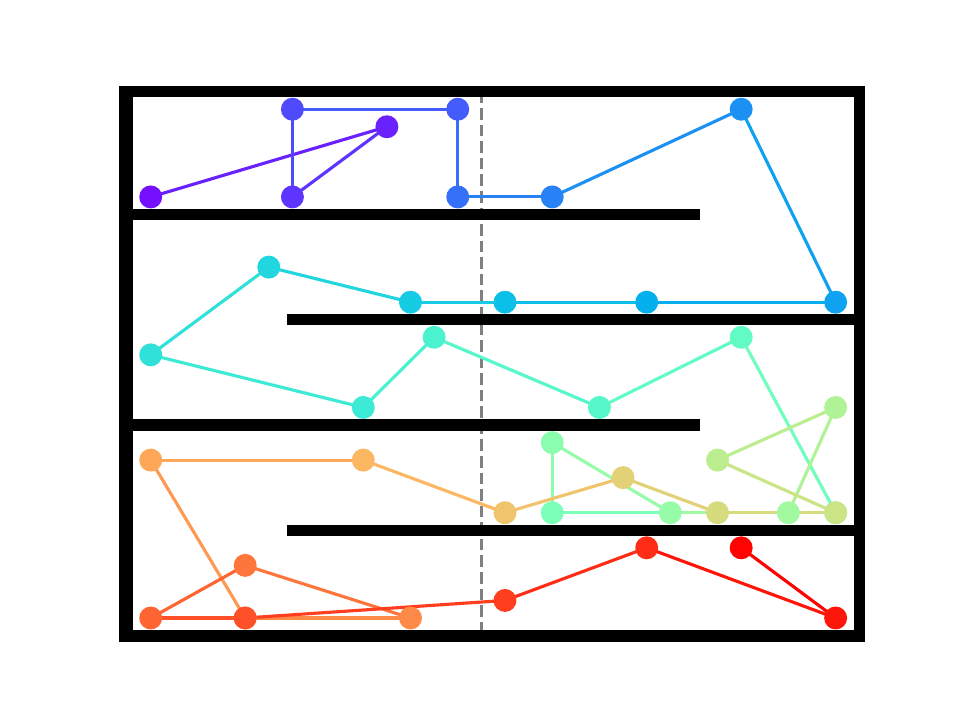}}
\caption{
 The black bars represent the walls and the dotted line represents the danger zone in the 2d maze.
 $40$ skills discovered by our method are shown.
 For each skill, the circles represent the average position of locations visited often.
 Colors change from early skills in purple to late skills in red. 
 Consecutive reward functions are connected by a line.
 The agent learns increasingly complex skills, until it reaches the bottom right corner of the maze; the hardest part to reach in the whole maze. }
\label{fig:forward_original}
\vspace{5.6cm}
\end{wrapfigure} 

The task consists of a 32 by 32 maze with several walls and a `danger zone'. 
The observation is given as a $32$x$32$x$1$ image with all values set to $0$, except a $1$ in the current position of the agent. 
The agent starts in the top left corner and can perform $5$ actions, either move in one of the cardinal directions or stay in the current position. If the agent moves into a wall it stays at its current position instead. If the agent is in the `danger zone' and moves up, moves down or stays, the episode is terminated and the agent is moved back to the starting position.
Figure \ref{fig:forward_original} shows the layout of the maze.
The `danger zone' ensures that random exploration will not work to reach many parts of the environment and lets us easily test both the increasing complexity of the reward functions and the importance of forward transfer.
We computed the expected number of steps to reach the bottom right corner with a random walk using Dynamic Programming. In expectation, $7 \cdot 10^{27}$ episodes are needed to do so. 
This shows that this maze is difficult to navigate.

\begin{wrapfigure}{r}{0.4\columnwidth}
\vspace{-10.3cm}

\centerline{\includegraphics[width=0.35\columnwidth]{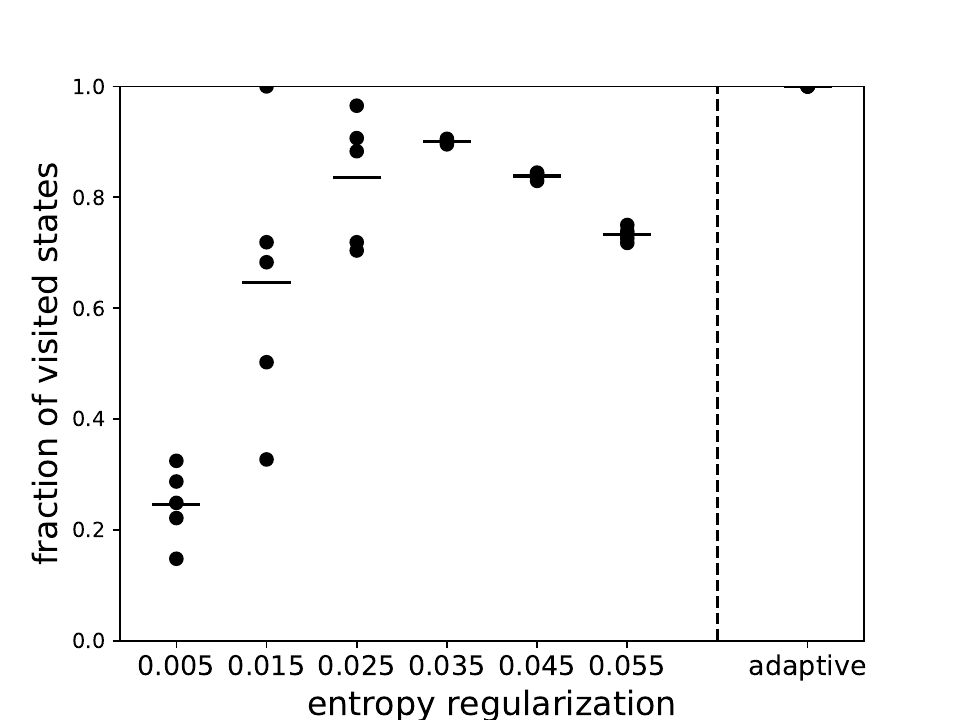}}
\caption{Total number of states visited by the learned policies after $60$ generations as a function of the entropy regularization used by the underlying A2C agent. The adaptive entropy model manages to visit all states in all runs.}
\label{fig:fixed_entropy}

\end{wrapfigure}

To train our agent we use the Advantage Actor Critic (A2C) \cite{mnih2016asynchronous} algorithm. To learn the rewards we use the full algorithm presented in Section \ref{sec:methods}.
We use the same architecture for the reward, policy and value networks, but do not share any parameters. The architecture is a ReLU network with 2 convolutional layers followed by 2 fully connected layers. 
For the exact hyper-parameters see Table \ref{table:2d-hp} in Appendix \ref{app:2d-hp}.
Figure \ref{fig:forward_original} plots the most visited locations for each skill of one run of our algorithm. 
The first few skills visit points near the origin, later skills start moving to harder to reach parts of the state space. After roughly $40$ iterations they reach the bottom right part. As stated before, this would take unreasonably long when using only random exploration.

Inspired by the BRAX library \cite{freeman2021brax}, we implemented both the environment and an A2C agent inside a single JAX \cite{jax2018github} compiled function.
By doing this, the computation graph of the environment and agent are optimized jointly and both run on the GPU. 
This eliminates the need to send data between the CPU and GPU which is one of the main bottlenecks in RL. 
Using just one NVIDIA RTX 3090 GPU,
the training process runs at over one million frames per second which enables training of agents in just a few seconds.
This allows us to experiment quickly and at a lower economic and environmental cost. 
We believe this code is useful for the RL community on its own and provide it in the supplementary material.

\subsubsection{Forward transfer of skills}
\label{sec:navigation_forward}

As pointed out in Section \ref{sec:method_transfer}, our reward functions become too complex to be learned from scratch with random exploration in a reasonable number of steps. 
When this happens, our agent must rely on transferring knowledge from previous generations.
Our navigation task is specifically designed to test this transfer ability, as random exploration would never reach the bottom right corner ($7\cdot10^{27}$ episodes in expectation).

\begin{wrapfigure}{R}{0.5\columnwidth}
\vspace{-.6in}
\centering
\begin{tabular}{ccc}
\subfigure[Fixed $0.0025$]{\includegraphics[width = .2\columnwidth]{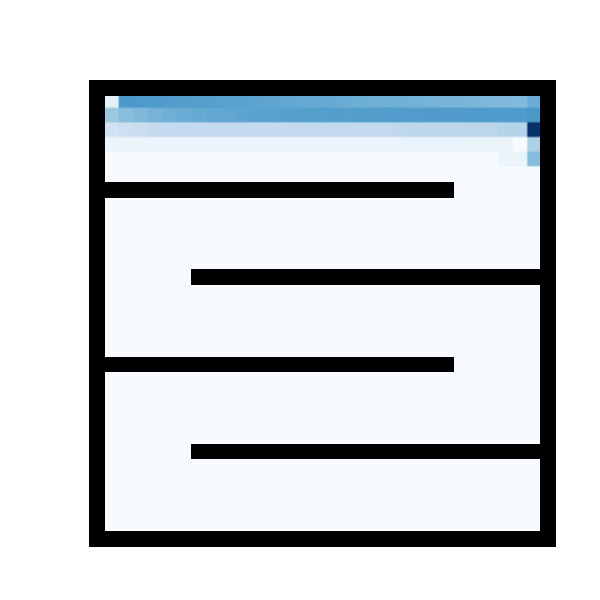}} &
\subfigure[Fixed $0.035$]{\includegraphics[width = .2\columnwidth]{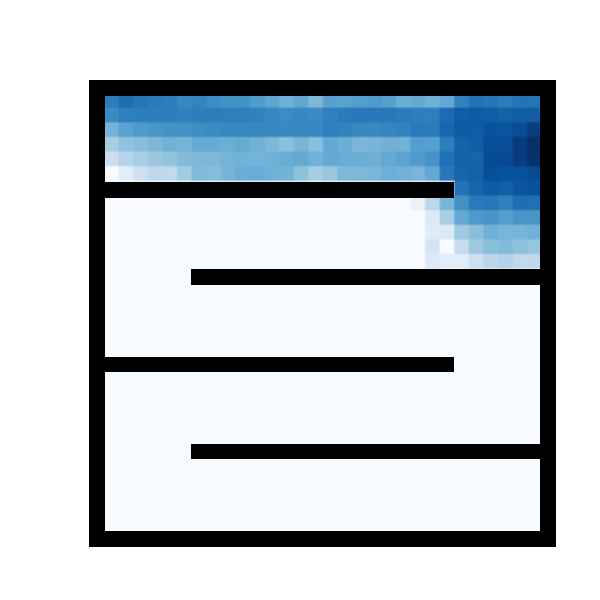}} & \vspace{-2em}\\

\subfigure[Adaptive]{\includegraphics[width = .2\columnwidth]{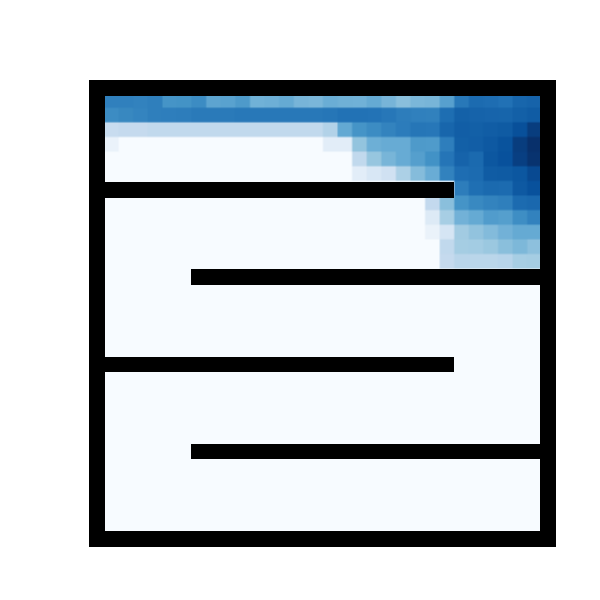}}  &
\subfigure[Reward Function]{\includegraphics[width = .2\columnwidth]{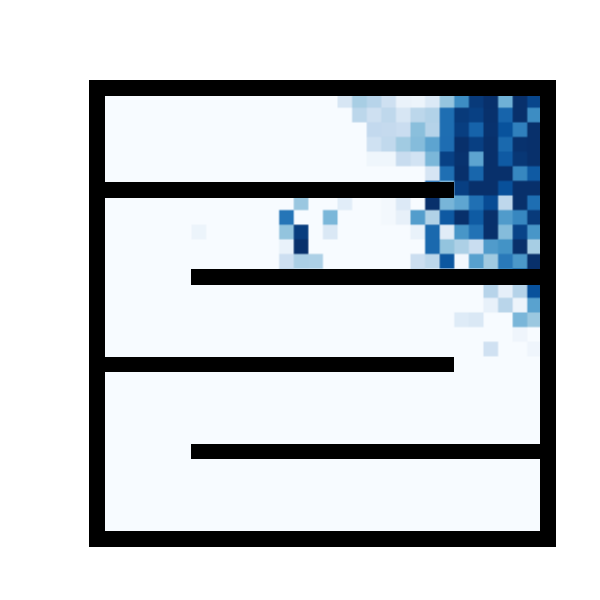}} 
\end{tabular}
\caption{States visited by policies trained to solve a reward function (d) that was created in the training process of Figure \ref{fig:forward_original}. We trained two policies with a fixed entropy regularization of $0.0025$ (a) and $0.035$ (b). (c) shows an agent trained with an entropy regularization of $0.0025$ in non-rewarding states and of $0.035$ in rewarding ones.}
\label{fig:adaptive_entropy}
\vspace{-.2in}
\end{wrapfigure}

We experimentally evaluate the three forward transfer mechanisms proposed in Section \ref{sec:method_transfer}: \textbf{Value reuse}, \textbf{Policy feature reuse} and \textbf{Guiding policy}.
The reward functions from Figure \ref{fig:forward_original} serve as tasks, sorted according to creation order. We train ablations of the three mechanisms sequentially on these tasks.
This allows us to ignore the skill discovery process and only measure the forward transfer of skills. 
We repeat each experiment three times. The agent with all mechanisms always manages to solve\footnote{We consider a reward function as solved if the agent manages to find positive reward.} all reward functions.
On the other hand, the Policy, Value and Guiding ablations fail to learn after solving $29.3 \pm 9.5$, $18.7 \pm 11.6$ and $19 \pm 8.5$ reward functions, respectively.

\subsubsection{Speeding-up deep exploration}
\label{sec:deep-exploration}

One key parameter when training actor critic methods is entropy regularization. 
In our method, policies with a lot of entropy generate a diverse set of negative samples. Diverse negative samples lead to reward functions that evolve more in each generation.
This is especially beneficial in environments where many steps are necessary to reach certain states, like in this 2D navigation task or Montezuma's Revenge.
We empirically verify this claim by training a set of agents with varying levels of entropy regularization.
Figure \ref{fig:fixed_entropy} shows the coverage of the state space after $60$ generations as a function of the entropy regularization.
We observed that higher entropy leads to a faster coverage of the state space up to a certain threshold. However, too much entropy leads to policies that do not learn to reach the rewarding states when these are far away from the origin. 
Observe that this problem arises independent of the training procedure as entropy regularization changes which policy is considered optimal.

Hence, we want to have policies that gather diverse samples to change the reward function but can also be deterministic in parts of the trajectory that already have no reward. To achieve this, we use adaptive entropy regularization. That is, we apply a small entropy regularization term everywhere. We increase the entropy regularization in states where the reward function is positive, i.e. states in which we want to decrease the reward function. 
See Figure \ref{fig:adaptive_entropy} for a visualization of the impact of the different entropy regularization strategies.
With this, the method manages to consistently visit all states in the maze.  This technique was necessary to create the reward functions from Figure \ref{fig:forward_original}.

\subsection{Robotic Environments}
\label{sec:robotics}

Quantitatively measuring unsupervised reinforcement learning progress remains an open problem and active area of research~\cite{gu2021braxlines}. 
Because of this, it is still important to visualize and qualitatively study the learned skills.
We do this in Section \ref{sec:robo_qual}. In the next two Sections, \ref{sec:robo_zero} and \ref{sec:robo_MI}, we quantitatively measure the performance of our algorithm on downstream task performance and with the so-called particle-based mutual information metric. 

In contrast to the previous experiments, the action space is continuous and the input modality is now a vector of features, like relative position, angle and speed of the different joints.
We do not use adaptive entropy here, as we need deterministic policies to maximize the downstream task performance and the mutual information metric.
Given that the environments do not have far away regions to reach, we did not implement the guiding algorithm here.
We do ablations for the other two transfer mechanisms.
Note that the agent does not see the $x$-$y$ position. Details about hyper-parameters and architecture can be found in Appendix~\ref{app:robo-arc}.

 In addition to skill discovery methods, we also compare our methods to RND~\cite{burda2018exploration} and `Disagreement'~\cite{pathak2019self}, two of the most prominent works in intrinsic motivation. By default, these methods only create a single final policy. We extract multiple skills by taking checkpoints of the policy  regularly during training 
\footnote{We thank anonymous reviewers `Wvtf' and `uUdP' for suggesting this baseline.}. To our knowledge, this has not been done before and we believe this showcases how closely related intrinsic motivation and unsupervised skill discovery are. We tuned the hyper-parameters of both methods extensively to maximize the Mutual Information metric (c.f. Section~\ref{sec:robo_MI}).

\subsubsection{Qualitative Analysis}\label{sec:robo_qual}

\begin{wrapfigure}{r}{0.6\columnwidth}
\vspace{0cm}
\centerline{\includegraphics[width=0.5\columnwidth]{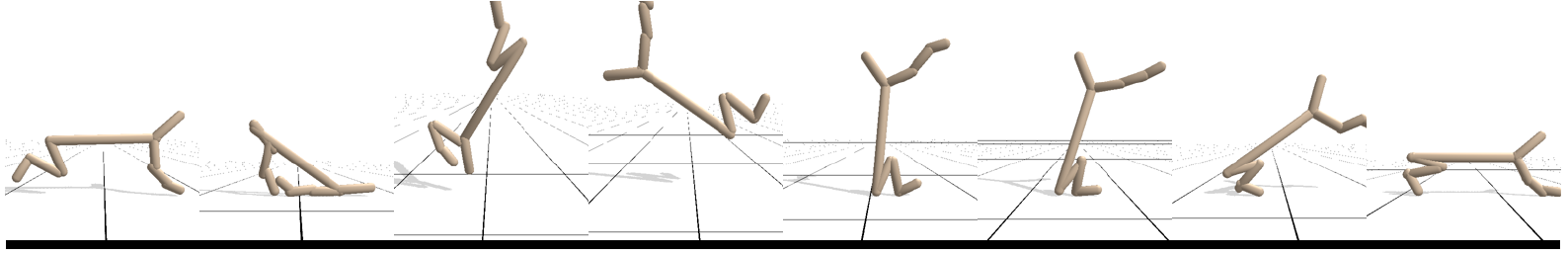}}
\centerline{\includegraphics[width=0.5\columnwidth]{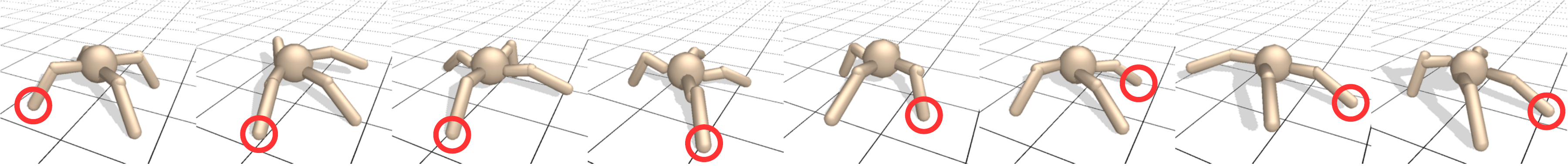}}
\centerline{\includegraphics[width=0.5\columnwidth]{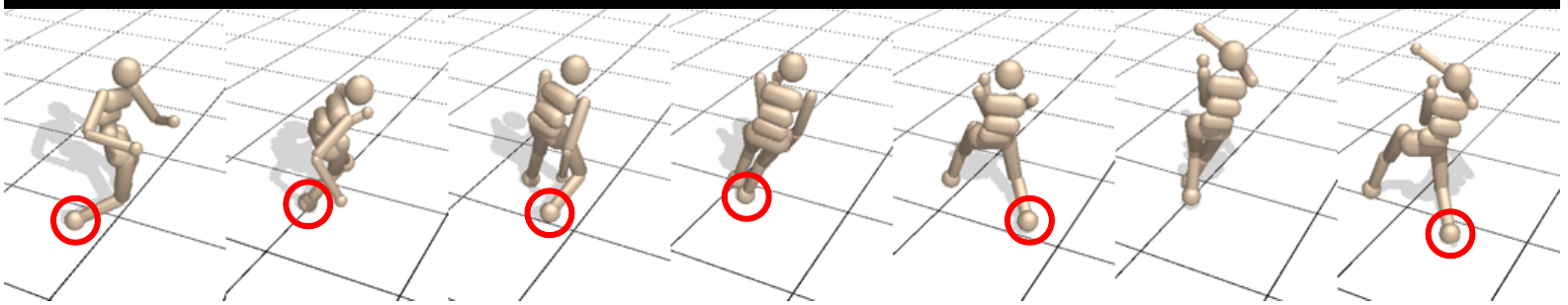}}
\centerline{\includegraphics[width=0.5\columnwidth]{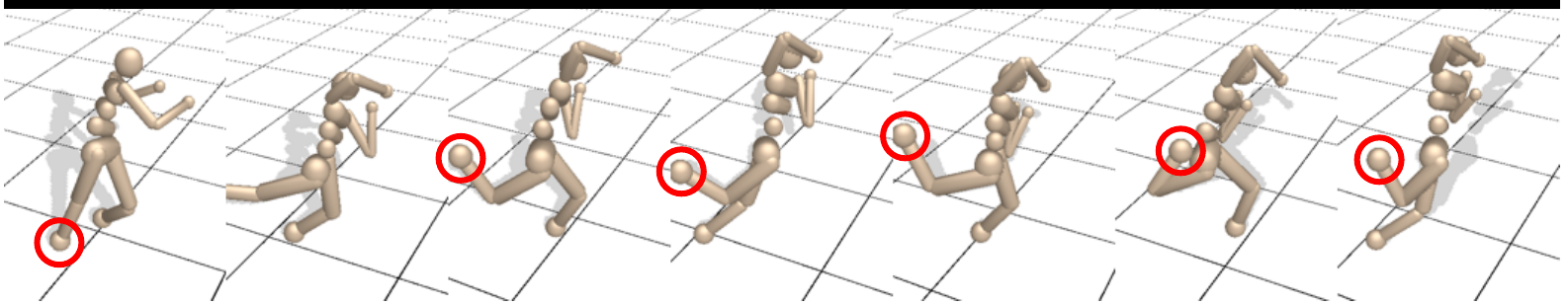}}
\caption{Visualization of four learned robotic skills. One of the legs is tracked with a red circle when it is visible. The \textsc{Half-Cheetah} learns to do a front-flip. The \ant does a partial rotation around its torso and then starts running. The upper \hum also does a partial rotation and then runs backwards. The lower \hum jumps on one leg while moving forward. }
\label{fig:robo-visual}
\vspace{2cm}
\end{wrapfigure}

In Figure \ref{fig:robo-visual} we illustrate a selection of particularly interesting skills. Videos of these and additional skills can be found \href{https://github.com/amujika/Open-Ended-Reinforcement-Learning-with-Neural-Reward-Functions/tree/main/videos}{\color{blue}{here}}. In Figure \ref{fig:robo-skills} we see how the velocity of learned skills in \ant and \hum evolve over training. See Appendix \ref{app:robo-fig} for the other runs.
One can see that consecutive skills slightly change the direction and speed of moving.
However, it is important to realize that the observations contain tens to hundreds of dimensions.
Thus, the skills can encode much more complex behaviors than speed and direction of movement.
For example, Figure \ref{fig:robo-visual} shows that the skill involves keeping one leg in the air on top of moving in the right direction.

 We also show scatter plots for RND and `Disagreement' in Appendix~\ref{app:robo-fig}. While the two methods also work well in the \hum environment (c.f. Figures~\ref{fig:rnd-hum-xy} and \ref{fig:dis-hum-xy}), they do not find any useful behavior in the \ant environment (c.f. Figures~\ref{fig:rnd-ant-xy} and \ref{fig:dis-ant-xy}). These approaches reward all novel states, they lead to policies that cover large regions of the state space, as can be seen in \textsc{Ant}.
We believe that the inherent instability of \hum forces the policies to be more specific.

\subsubsection{Zero-Shot Transfer}\label{sec:robo_zero}

\begin{wrapfigure}{r}{0.6\columnwidth}
\vspace{-4.5cm}
\includegraphics[width=0.6\columnwidth]{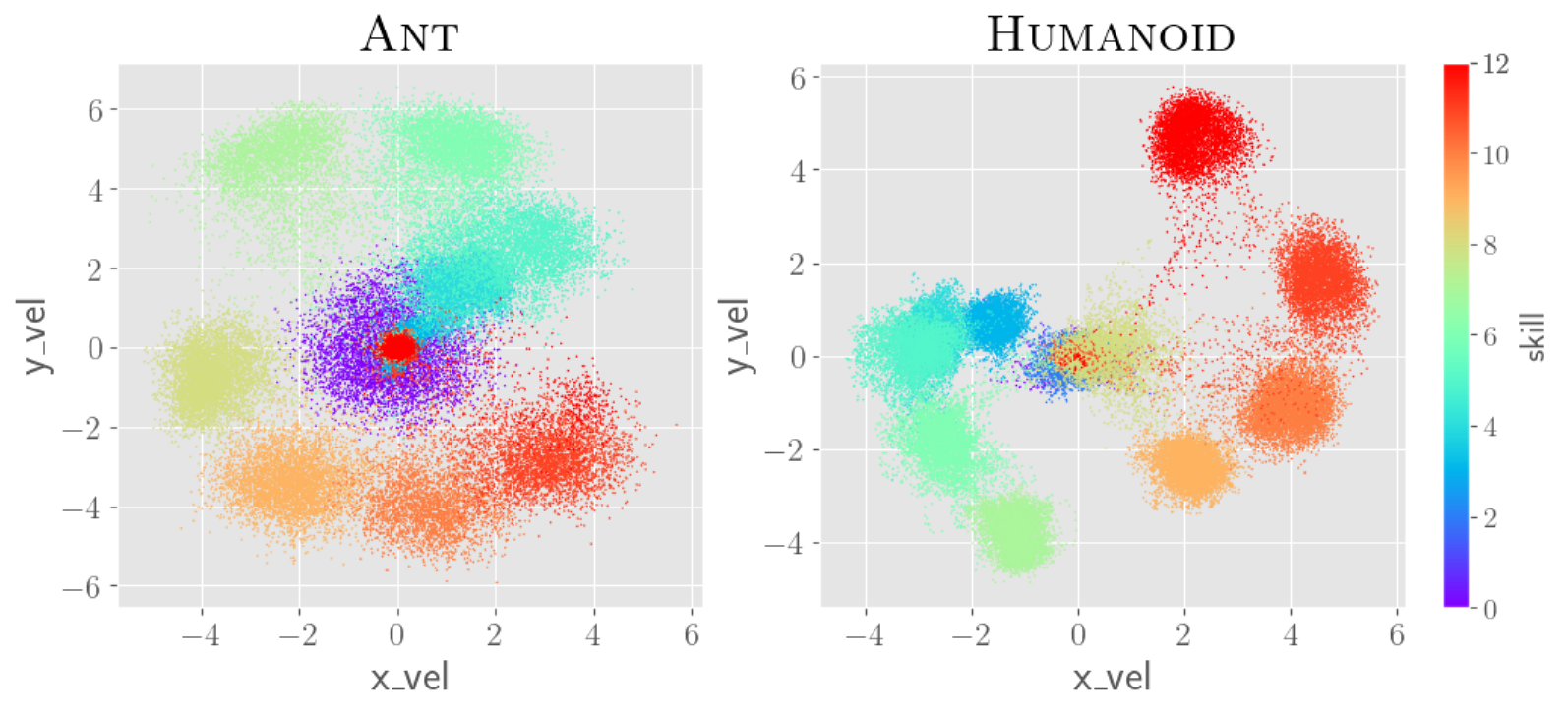}
\caption{Scatter-plots of the $x$- and $y$-velocity of the states visited by the first $13$ skills of one run in \ant and \textsc{Humanoid}.  
The second half of the trajectory is shown. 
Colors change from early skills in purple to late skills in red.}
\label{fig:robo-skills}
\vspace{-0.2cm}
\end{wrapfigure}

After discovering $50$ unsupervised skills, we identify the skill that aligns best with the reward given by the environment. We report this zero-shot\footnote{Technically, we query the true reward function to identify the skills, but do not perform any training with it.}
performance in Table \ref{table:zero-shot}.
As a baseline, we train an agent from scratch and measure how long it takes it to reach an equivalent performance. We use the Optuna hyper-parameter optimizer \cite{akiba2019optuna} to find hyper-parameters which maximize the speed of learning beforehand. The exact procedure can be found in Appendix~\ref{app:robo-zero}.

The simplest environment, \textsc{Half-Cheetah}, benefits the least from unsupervised learning, while the hardest one, \textsc{Humanoid}, benefits the most. 
We report the zero-shot performance of RND, `Disagreement' and the ablations in Table~\ref{table:app-zero-shot} in the Appendix. 
As already seen in Section~\ref{sec:robo_qual}, the intrinsic motivation methods do not find any interesting behavior in \textsc{Ant}.
Conversely, they perform well on the \hum environment, the latter one even slightly outperforming our method\footnote{Note that the tuning RND and `Disagreement' for the $x$-velocity MI-metric promotes locomotion.}.

\begin{wraptable}{r}{7cm}
\vspace{-.45cm}
\caption{Zero-shot environment reward of our algorithm and the number of steps a supervised PPO agents needs to match it. Both columns averaged over $10$ repetitions.}
\label{table:zero-shot}
  \begin{tabular}{lcc}
\toprule
Task & Zero-shot & Steps from \\
 & reward & scratch\\
\midrule
Cheetah & $1094 \pm 1130$ & $340K \pm 50K$ \\
Ant          & $2506 \pm 511$ & $1.2M \pm 0.24M$\\
Humanoid     & $9092 \pm 1063$ & $55M \pm 27M$ \\

\bottomrule
\end{tabular}
\end{wraptable}

\subsubsection{Particle-based Mutual Information metric}\label{sec:robo_MI}
Measuring how well the agent can control relevant state dimensions is another way to track progress in Unsupervised reinforcement learning.
This is measured using the mutual information between state dimensions and skills.
While high-dimensional estimation of mutual information is an active area of research, sampling can be an effective form of estimation in the 1-dimensional case,  c.f. Algorithm 1 in \cite{gu2021braxlines}.
We report the particle-based mutual information for the $x$-velocity in Table \ref{table:MI}, using the same bucketing strategy\footnote{We split the dimension in $1000$ bins in the $[-10,10]$ range. As in \cite{gu2021braxlines}, we only take the second half of each trajectory. The initial state is independent of the skill and thus the beginning of the trajectory does not tell anything about the gained controllability.} as in \cite{gu2021braxlines}.

Our method heavily outperforms \textsc{DIAYN} and \textsc{DADS}, when both methods look at the complete observation space.
\diayn and \textsc{DADS} achieve diversity by learning a set of skills that can be correctly labeled by a neural network.
In high-dimensional spaces this is easy to do by relying on a small subset of all state dimensions.
This leads to non-diverse behaviors across most state dimensions.
Even when expert knowledge about relevant dimensions is supplied to other methods, i.e. only taking the $x$- and $y$-velocity into account, our method still fares well. Particularly, in the most complex environment, \textsc{Humanoid}, our method performs best.
We believe that the iterative increase in complexity leads to better coverage of hard to reach regions of the state like high speed or running backwards.
With all this, our approach achieves greater controllability of the $x$-velocity than \diayn and \textsc{DADS} without any kind of feature engineering.

\begin{wraptable}{r}{8.8cm}
\vspace{-0.4cm}
 \caption{Particle-based mutual information metric for the $x$-velocity. Results are averaged over $10$ runs. Algorithms with feature engineering only consider $x$-$y$ velocities. Baselines taken from \cite{gu2021braxlines}}
\label{table:MI}
 \begin{tabular}{lccr}
\toprule
Task & Method & Feature & MI$(s,z)$ \\
 & & Engineering & \\
\midrule
Cheetah & Ours  &   \ding{55} &$1.40 \pm  0.21$\\
Cheetah & \diayn &  \ding{55} &$0.49 \pm 0.16$  \\
Cheetah & $\textbf{DIAYN}_p$ & \checkmark  &$\mathbf{1.82 \pm 0.20}$ \\
Cheetah & \gcrl & \checkmark &$1.63 \pm 0.16$ \\
\midrule
Ant          & \textbf{Ours}   &  \ding{55} &$\mathbf{1.33 \pm 0.11}$\\
Ant          & RND &  \ding{55} &$0.26 \pm 0.15$\\
Ant          & Disagreement &  \ding{55} &$0.08 \pm 0.03$\\
Ant          & \diayn &  \ding{55} &$0.07 \pm 0.01$ \\
Ant          & DADS &  \ding{55} & $0.32 \pm 0.06$ \\
Ant          & $\textsc{DIAYN}_p$  & \checkmark &$1.12 \pm 0.27$\\
Ant          & \gcrl & \checkmark  &$1.22 \pm 0.19$\\
\midrule
Humanoid     & \textbf{Ours}  &  \ding{55} &$\mathbf{1.29 \pm 0.25}$ \\
Humanoid     &  RND &  \ding{55} &$0.94 \pm 0.17$ \\
Humanoid     & Disagreement &  \ding{55} &$1.05 \pm 0.14$ \\
Humanoid     & \diayn &  \ding{55} &$0.07 \pm 0.01$ \\
Humanoid     & DADS &  \ding{55} & $0.24 \pm 0.06$ \\
Humanoid     & $\textsc{DIAYN}_p$  & \checkmark &$0.93 \pm 0.13$ \\
Humanoid     & \gcrl & \checkmark &$0.77 \pm 0.15$ \\

\bottomrule
\end{tabular}
\vspace{-1cm}
\end{wraptable}

Surprisingly, our novel\footnote{The approaches are not new, but using them to generate checkpoints that can be used as skills has not been done before to the best of our knowledge.} baselines, RND and `Disagreement', also achieve better performance than \diayn and \textsc{DADS} in the \hum environment, but not in \textsc{Ant}. As already discussed in Section~\ref{sec:robo_qual}, we believe that the larger instability of \hum compared to \ant is the reason for this difference in performance.

\subsection{Montezuma's Revenge}
\label{sec:montezuma}

To show the generality of our approach we evaluate it on the notoriously hard Montezuma's Revenge Atari game. In this game, the agent controls a character in a complex 2d world with several rooms. Appendix \ref{app:montezuma-env} shows the initial room and various items with which the agent may interact. Same as in \cite{mnih2015human}, the observation is a stack of the last $4$ frames. This gives the agent information about speed and direction of movement. This is done for all networks, that is, the value, policy and neural reward networks. We use the same simple CNN architecture as in \cite{mnih2015human} for all three networks. For exact learning details see Appendix \ref{app:montezuma-exp}.

Our algorithm uses finite episode lengths because once it reaches a rewarding state, the agent can stay there forever.
Because of this, we reset the environment every $500$ steps.

One of the main difficulties when dealing with Montezuma's Revenge is that it cannot be simulated as fast as the other studied environments. On top of that, an agent can learn hundreds of \textit{different} skills without ever leaving the first room. 
Finally, skills learned by our agent evolve from simple to very complex and extended in time.
In the beginning, the agent just needs to stay close to the initial position.
By the end of training, the agent learns to collect a key, open the door, avoid several enemies and visit four different rooms.
In the most complex skills, it takes the agent several hundred steps to reach a rewarding state.
This means that experiments can take several days to visit a different room. 
 
\begin{wrapfigure}{r}{0.5\columnwidth}
\vspace{-.35cm}
\begin{center}
\subfigure[]{\includegraphics[width=.2\columnwidth]{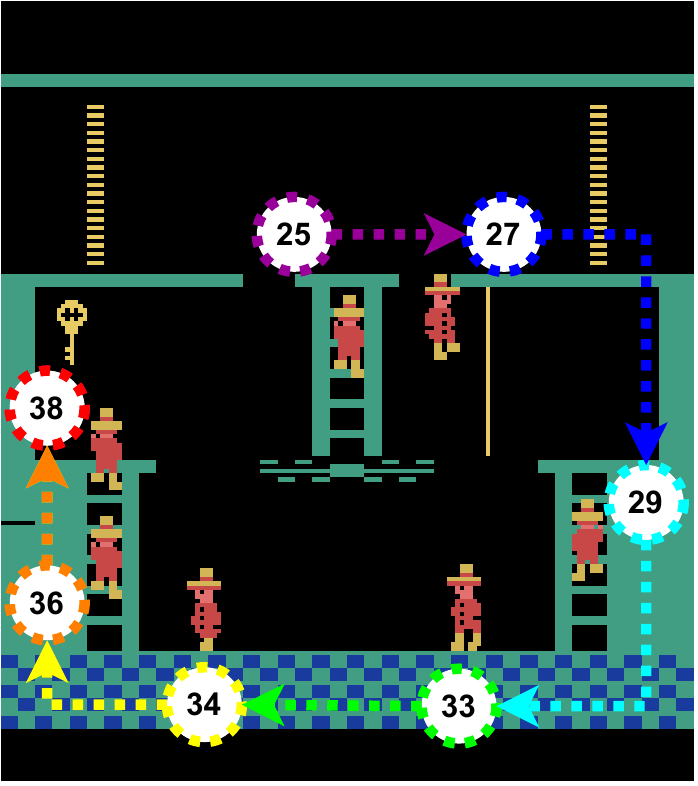}\label{fig:mz-skills}}
\subfigure[]{\includegraphics[width=.24\columnwidth]{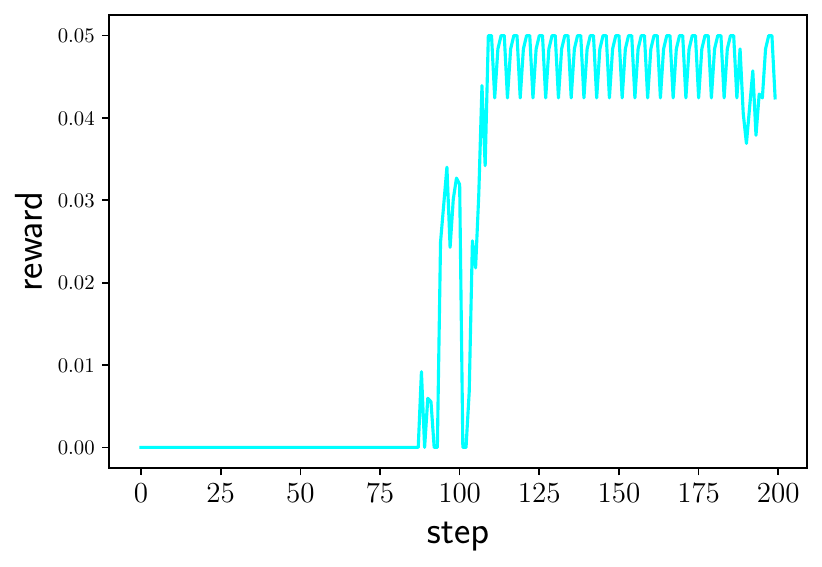}\label{fig:mz-rewards}}
\caption{(a) The position of the agent in the most rewarding state of each skill. The skills are numbered by the creation order. (b) The rewards received by skill $29$ during part of its trajectory. This shows the long term nature of the discovered skills.}
\label{fig:montezuma}
\end{center}
\vspace{-1cm}
\end{wrapfigure}

In order to save computation, we adapt the number of training steps to ensure the agent has learned each skill. See Appendix \ref{app:montezuma-episode} for details.

We found out that the first thing the agent learns is to get the life counter to $0$. Then it starts exploring the room, with $0$ lives already.
This is because losing a life happens very easily during random exploration and because once the agent reaches $0$ lives, all future positive samples will have $0$ lives.
This has two side effects. On the one hand, the initial $100-200$ steps of an episode are spent losing lives. On the other hand, exploration is harder, as the agent will reach a terminal state very easily.
Because of this, we cut out the part of the image that shows the remaining lives, see Appendix \ref{app:montezuma-env}.

Figure \ref{fig:mz-skills} visualizes different skills. The first observations which get the maximum possible reward are overlaid. All the shown skills go to the bottom and kill the skull first. Then, they move to a specific location. It can be seen that later skills explore states further and further away from the initial state.
Figure \ref{fig:mz-rewards} shows that the skill receives no reward for many steps. Only after the agent has killed the skull being on the ladder becomes rewarding\footnote{Using the ladder is needed to kill the skull, but the reward function gives no reward when the agent is on the ladder before killing the skull.}.
This shows that both the reward network and the agent have learned a lot of the concepts that are necessary to tackle Montezuma's Revenge. This includes controlling the agent in the 2d environment, killing enemies and interacting with different objects. All of this without ever accessing the original reward function.
Eventually, after learning a few hundred skills, the neural reward function pushes the rewarding states all the way to the fourth room.
\href{https://github.com/amujika/Open-Ended-Reinforcement-Learning-with-Neural-Reward-Functions/blob/main/videos/montezuma_other_rooms.gif}{\color{blue}{This video}} shows a run of this skill.

\section{Limitations}
Our approach has three main limitations: the randomized exploration strategy, the re-learning necessary for each new skill and the depth-first nature of the search of skills.

We have shown that the random exploration allows for discovering diverse skills. However, it can lead to slow progress in complex regions of the state space. Using more sophisticated exploration policies could lead to faster evolution of the skills. A potential avenue for improving this is using previously discovered skills as exploration policies.

As stated in Section \ref{sec:method_transfer}, we use multiple techniques to make the learning of the next policy more efficient. Still, the vast majority of our compute is used for learning these policies. Using more advanced transfer learning or even meta-learning algorithms, could make the policy learning more efficient. Another possibility would be to relabel previously seen trajectories with ideas similar to \cite{rauber2017hindsight}.

Finally, our approach only uses the current skill to create the next one. 
On a high level, it uses a single search point at each time. Then, the iterative increase in complexity of the skills leads to a DFS-like exploration.
This may not always be optimal. For example, in Montezuma's Revenge it causes the agent to explore only one of the two possible exits of the first room.
On top of that, our algorithm can sometimes get stuck in places were backtracking is complicated or impossible, see Figure \ref{fig:2d-t} in the Appendix.
This also happens in Montezuma's Revenge when the life-counter is not removed.
Figuring out smart back-tracking strategies or keeping multiple search points could address this problem and make our algorithm even more efficient.
\section{Conclusion}
We have presented an unsupervised reinforcement learning algorithm that uses reward functions encoded by neural networks. Our algorithm alternates between increasing the complexity of the reward function and transferring previous knowledge to learn a new skill that finds rewarding states.
This allows it to learn an unbounded number of skills.

We have thoroughly tested the different components of our model in a 2d navigation task. 
This has allowed us to better understand our method in practice. 
We have shown that our method works both with high dimensional feature inputs, in robotic environments, and pixel inputs, in Montezuma's Revenge. Our algorithm learned a diverse set of skills in both settings. 
In \textsc{Humanoid} and Montezuma's Revenge, skills found by our method achieve a zero-shot performance that takes millions of steps to learn in the classical reinforcement learning setup.

We believe our algorithm is one step in a direction that may one day allow reinforcement learning agents to fully understand an environment without making use of any predefined reward function.
Just like in Computer Vision and Natural Language Processing, this will lead to agents that need very few labels from the task at hand to be able to solve it and will drastically expand the applicability of reinforcement learning. 

\section*{Acknowledgements}
 We would like to thank Xun Zuo for helping with the implementation of the maze environment, on top of many great discussions.
We would also like to thank Frederik Benzing and Yassir Akram for many fruitful discussions. 
Finally, we also thank the anonymous reviewers for the thorough and insightful reviews and discussions.
Robert Meier and Asier Mujika were supported by grant no. CRSII5\_173721 of the Swiss National Science Foundation.

\bibliography{bibliography}
\bibliographystyle{iclr2022_conference}

\newpage
\section*{Checklist}

\begin{enumerate}

\item For all authors...
\begin{enumerate}
  \item Do the main claims made in the abstract and introduction accurately reflect the paper's contributions and scope?
    \answerYes{}
  \item Did you describe the limitations of your work?
    \answerYes{} We discuss limitations and future directions of research in the Conclusion
  \item Did you discuss any potential negative societal impacts of your work?
    \answerNo{} We developed a new approach for open-ended learning. While we hope that our work moves the understanding of unsupervised open-ended processes forward in the long term, we do not expect any immediate societal impacts in our work.
  \item Have you read the ethics review guidelines and ensured that your paper conforms to them?
    \answerYes{}
\end{enumerate}

\item If you are including theoretical results...
\begin{enumerate}
  \item Did you state the full set of assumptions of all theoretical results?
    \answerNA{} No theory
        \item Did you include complete proofs of all theoretical results?
    \answerNA{} No theory
\end{enumerate}

\item If you ran experiments...
\begin{enumerate}
  \item Did you include the code, data, and instructions needed to reproduce the main experimental results (either in the supplemental material or as a URL)?
    \answerYes{}
  \item Did you specify all the training details (e.g., data splits, hyperparameters, how they were chosen)?
    \answerYes{}
        \item Did you report error bars (e.g., with respect to the random seed after running experiments multiple times)?
    \answerYes{}
        \item Did you include the total amount of compute and the type of resources used (e.g., type of GPUs, internal cluster, or cloud provider)?
    \answerYes{}
\end{enumerate}

\item If you are using existing assets (e.g., code, data, models) or curating/releasing new assets...
\begin{enumerate}
  \item If your work uses existing assets, did you cite the creators?
    \answerYes{}
  \item Did you mention the license of the assets?
    \answerYes{}
  \item Did you include any new assets either in the supplemental material or as a URL?
    \answerYes{} Included code, we will release it under the Apache 2 license or similar
  \item Did you discuss whether and how consent was obtained from people whose data you're using/curating?
    \answerNA{}
  \item Did you discuss whether the data you are using/curating contains personally identifiable information or offensive content?
    \answerNA{}
\end{enumerate}

\item If you used crowdsourcing or conducted research with human subjects...
\begin{enumerate}
  \item Did you include the full text of instructions given to participants and screenshots, if applicable?
    \answerNA{}
  \item Did you describe any potential participant risks, with links to Institutional Review Board (IRB) approvals, if applicable?
    \answerNA{}
  \item Did you include the estimated hourly wage paid to participants and the total amount spent on participant compensation?
    \answerNA{}
\end{enumerate}

\end{enumerate}

\appendix
\section{Pseudo-Code}\label{app:pseudo}

      \begin{algorithm}[H]
        \caption{Open-Ended Neural Reward Functions}\label{alg:method}
        \begin{algorithmic}
   \STATE Initialize $\theta_{0}$ and $\psi_0$ and $O_{neg\_all}$. 
   \STATE $i \leftarrow 0$
   \WHILE{\textsc{TRUE}}
   \STATE $O_{neg_i}$, $O_{pos_i} \leftarrow \emptyset , \emptyset$
   \STATE Set $R_{\psi_{i}}$ as the reward of the MDP
   \STATE Train $\pi_{\theta_i}$ and $V_{\theta_i}$ using an actor-critic algorithm 

   \FOR{$j=0$ {\bfseries to} $b$}
   \STATE Reset the MDP to the initial state
   \STATE Follow $\pi_{\theta_i}$ for $k$ steps and add the observed states to $O_{neg}$
   \STATE Follow a random policy for $k'$ steps and add the observed states to $O_{pos}$ 
   \ENDFOR
   \STATE $O_{neg\_all_i} \leftarrow O_{neg\_all_{i-1}} \cup O_{neg_i}$

   \STATE $\psi_{i + 1} \leftarrow \psi_{i}$
   \STATE Train ${\psi_{i+1}}$ with $\mathcal{L_\psi}$, $O_{pos_i}$, $O_{neg_i}$ and $O_{neg\_all_i}$
   \STATE $\theta_{i + 1} \leftarrow \theta_{i}$
   \STATE Set last layer of $\pi_{\theta_{i + 1}}$ to $0$
   \STATE $i \leftarrow i + 1$
   \ENDWHILE
        \end{algorithmic}
      \end{algorithm}

\section{2d Navigation: Experimental details}\label{app:2d-hp}

The episode length in the 2d navigation task is $250$ steps. The guiding phase lasts for $2/3$ of the total steps of the 1-steps A2C agent learning. We sample the guiding length uniformly at random in the $0$ to $200$ range.
For the negative samples we follow the learnt policy for $200$ steps. For the positive samples, we take random actions for $50$ steps after following the policy for $200$ steps.

\begin{table}[ht]
\caption{Hyperparameters for the 2d navigation task.}
\label{table:2d-hp}
\vskip 0.15in
\begin{center}
\begin{small}
\begin{sc}
\begin{tabular}{lc}
\toprule
Hyperparameter \\
\midrule
Base Entropy Regularization & $0.005$ \\
Extra Entropy Regularization & $0.05$ \\
Episode length & $250$ \\
A2C learning rate & $0.0001$ \\
A2C discount factor & $0.99$ \\
Batch-size & $2048$ \\
Steps per skill & $2048 \cdot 60000$\\
Positive/negative sample target value & $0.05/-0.05$ \\
\midrule
Reward network updates per skill & $500$\\
Reward network learning rate & $0.001$ \\
Reward network training batch size & $3 \cdot 256$ \\
\bottomrule
\end{tabular}
\end{sc}
\end{small}
\end{center}
\vskip -0.15in
\end{table}

\section{Robotic environments: Architecture and hyper-parameters}\label{app:robo-arc}
We use the PPO \cite{schulman2017proximal} implementation from \cite{freeman2021brax} with modifications such that it allows our training method to work.  In Table~\ref{table:ppo-hp} and Table~\ref{table:env-hp} we list the hyper-parameters we used. Note that we use a smaller learning rate for the bodies of the policy and value network. We found that this was beneficial for transferring more knowledge from the previous skill.

The environment specific parameters were found using Optuna \cite{akiba2019optuna}. For each environment we ran a search to optimize final performance on the environment rewards (running in positive $x$-direction). We used $200$ runs and trained each of them for the same number of steps as a skill in our method. This yielded hyper-parameters which are able to learn tasks in the corresponding environment. We did not take the downstream performance metrics (zero-shot performance and particle-based information) into account. Doing so would have exceeded our compute budget. In Table~\ref{table:nrf-robo} the architecture and hyper-parameters used for the neural reward functions are listed.  For the supervised reward training we use the Adam optimizer \cite{kingma2014adam}. All code can be found in the supplementary material.

For DADS~\cite{sharma2019dynamics} we used the code provided by the authors. For RND~\cite{burda2018exploration} and `Disagreement'~\cite{pathak2019self} we tuned the method specific hyper-parameters optimizing for the MI-metric of the $x$-velocity. In particular, for each environment and method we used $100$ runs of $200M$ environment steps per run.

\begin{table}[t]
\caption{Hyperparameters for PPO shared in all three environments.}
\label{table:ppo-hp}
\vskip 0.15in
\begin{center}
\begin{small}
\begin{sc}
\begin{tabular}{lc}
\toprule
BRAX PPO hyperparameters \\
\midrule
Policy hidden layer sizes & $[512, 512]$ \\
Value hidden layer sizes & $[512, 512]$ \\
Body learning rate multiplier & $0.5$ \\
Episode length & $1000$ \\
Action repeat & $1$ \\
Number of mini-batches & $32$ \\
Batch-size & $1024$ \\
Parallel environments & $2048$ \\
\bottomrule
\end{tabular}
\end{sc}
\end{small}
\end{center}
\vskip -0.15in
\end{table}

\begin{table}[t]
\caption{Environment specific PPO hyperparemeters.}
\label{table:env-hp}
\vskip 0.15in
\begin{center}
\begin{small}
\begin{sc}
\begin{tabular}{lccc}
\toprule
& \textsc{Half-Cheetah} & \ant & \hum \\
\midrule
Learning rate & $0.00010$ & $0.00029$ & $0.00017$ \\
Reward scaling & $0.24532$ & $5.58242$ & $0.15326$ \\
Unroll length & $3$ & $5$ & $6$ \\
Updates per epoch & $15$ & $14$ & $10$ \\
Discount factor & $0.99109$ & $0.92318$ & $0.99114$ \\
Entropy cost & $0.00062$ & $0.00200$ & $0.02087$ \\
Environment steps per skill & $10M$ & $10M$ & $20M$ \\
\bottomrule
\end{tabular}
\end{sc}
\end{small}
\end{center}
\vskip -0.15in
\end{table}

\begin{table}[t]
\caption{Neural reward function hyperparemeters for the BRAX environments.}
\label{table:nrf-robo}
\vskip 0.15in
\begin{center}
\begin{small}
\begin{sc}
\begin{tabular}{lc}
\toprule
Neural Reward Function hyperparemeters  \\
in BRAX environments\\
\midrule
Hidden layer sizes & $[87]$ \\
Hidden layer nonlinearity & tanh \\
Target value $a$ & $5$ \\
Gradient steps & $300$ \\
Learning rate & $0.001$ \\
Total batch size & $171$ \\
Negative steps & $300$ \\
Positive steps & $40$ \\
Number of sampling environments & $8192$ \\
Fraction of negative samples stored & $0.01$ \\
\bottomrule
\end{tabular}
\end{sc}
\end{small}
\end{center}
\vskip -0.15in
\end{table}
\section{Robotic environments: Zero-Shot Transfer}\label{app:robo-zero}

We use the Optuna hyperparameter tuning library \cite{akiba2019optuna} to create Table \ref{table:zero-shot}. We ran multiple optimization procedures. For each one, we fix a number of environment steps, then optimize the final performance on the environment reward over $200$ runs. We iteratively increased the number of timesteps until the highest score of the $200$ runs beats our averaged zero-shot performance. We take the hyperparemters of that run and train $10$ agents until they outperform our averaged zero-shot performance. The average number of training steps needed for this is reported in Table~\ref{table:zero-shot}. In the \hum environment three of the runs did not reach the score in $100M$ steps, at which point we stopped training. We nonetheless took $100M$ into the average.
The code for the hyperparameter search and the results of our conducted studies can be found in the supplementary material. 

\section{Robotic environments: Additional plots and Tables}\label{app:robo-fig}

\begin{figure}[ht]
\vskip 0.2in
\begin{center}
\subfigure[]{\includegraphics[width=.3\textwidth]{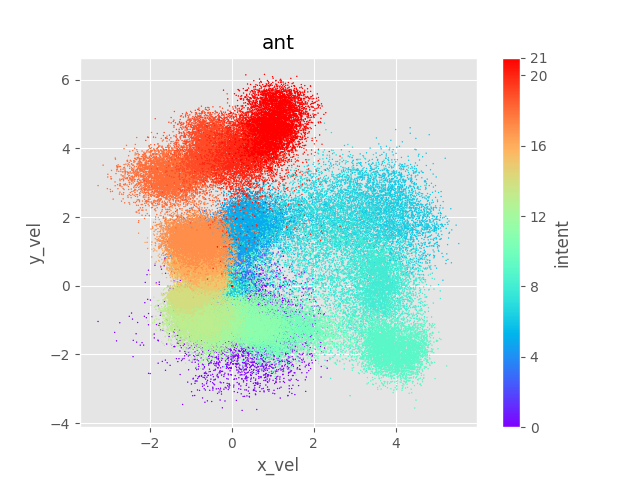}
}
\subfigure[]{\includegraphics[width=.3\textwidth]{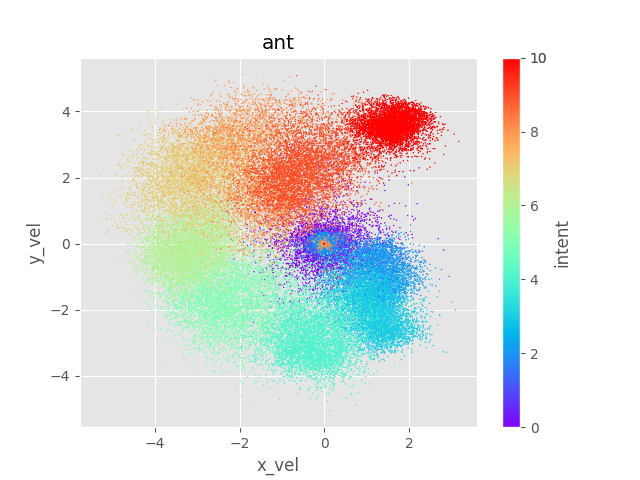}
}
\subfigure[]{\includegraphics[width=.3\textwidth]{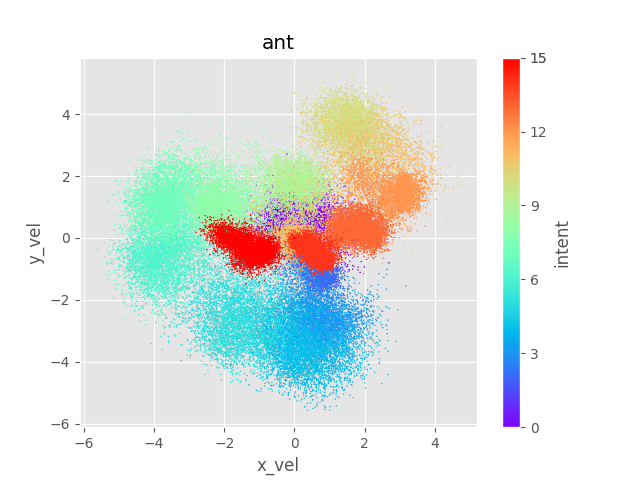}
}
\subfigure[]{\includegraphics[width=.3\textwidth]{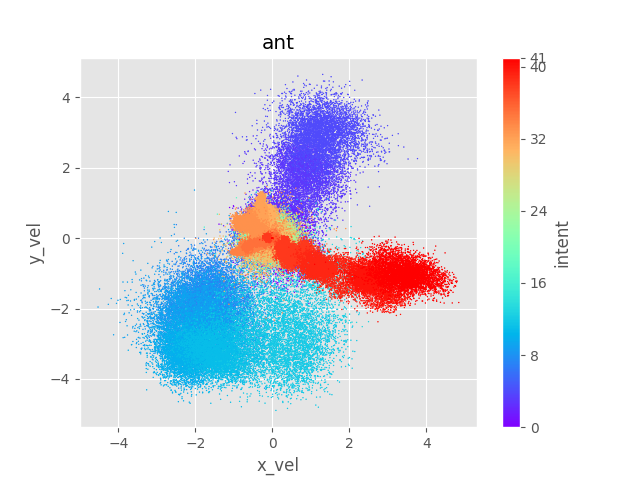}
}
\subfigure[]{\includegraphics[width=.3\textwidth]{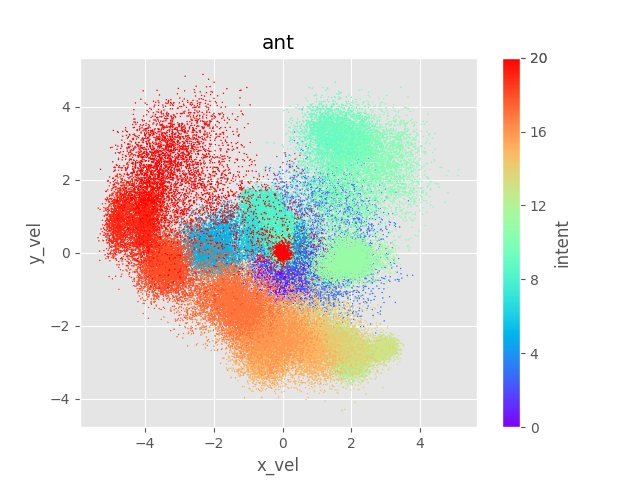}
}
\subfigure[]{\includegraphics[width=.3\textwidth]{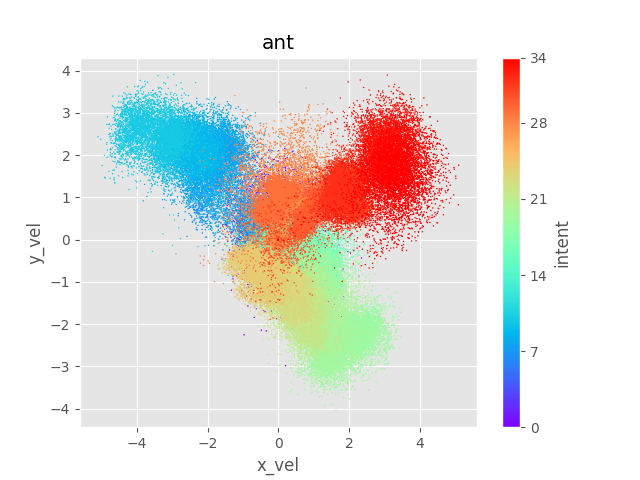}
}
\subfigure[]{\includegraphics[width=.3\textwidth]{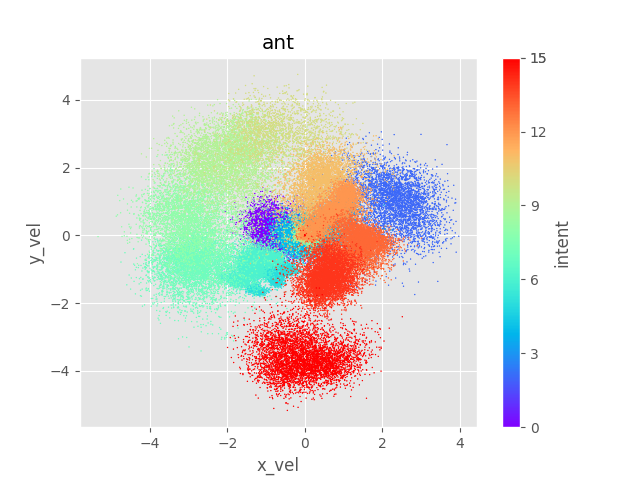}
}
\subfigure[]{\includegraphics[width=.3\textwidth]{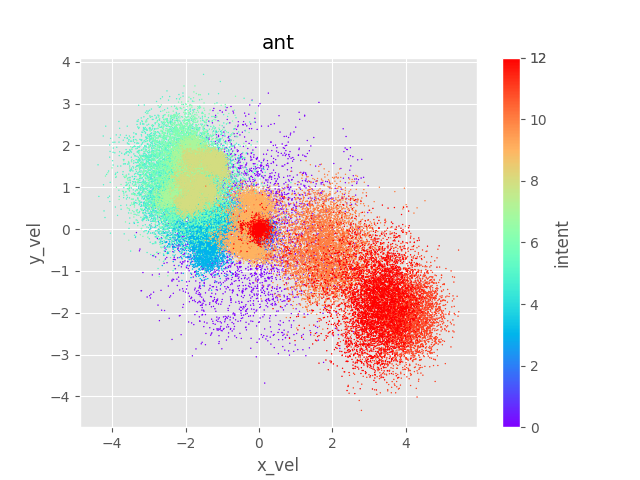}
}
\subfigure[]{\includegraphics[width=.3\textwidth]{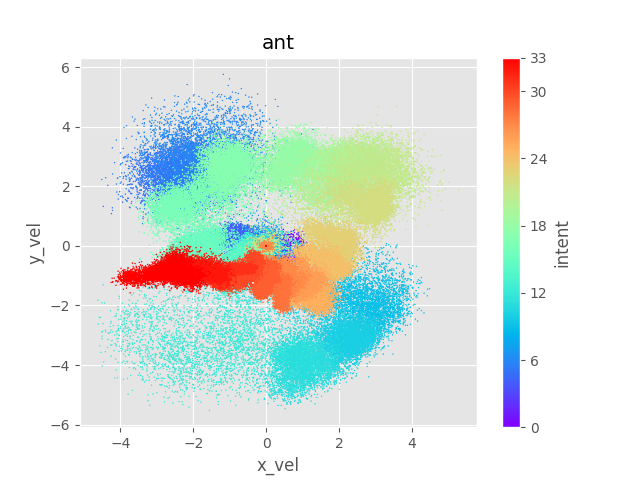}
}
\caption{Scatter-plots of the $x$- and $y$-velocity dimension of the states visited by the first $n$ skills of all additional $9$ runs in \ant. $n$ was picked by hand in each run. The second half of the trajectory is shown. 
Colors change from early skills in purple to late skills in red.  }
\label{fig:app-ant-xy}
\end{center}
\vskip -0.2in
\end{figure}

\begin{figure}[ht]
\vskip 0.2in
\begin{center}
\subfigure[]{\includegraphics[width=.3\textwidth]{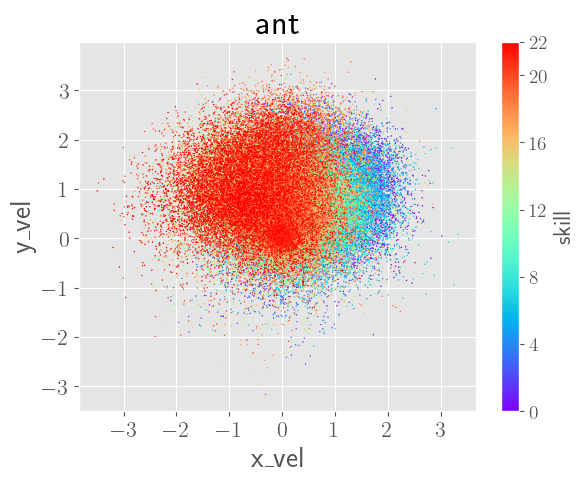}
}
\subfigure[]{\includegraphics[width=.3\textwidth]{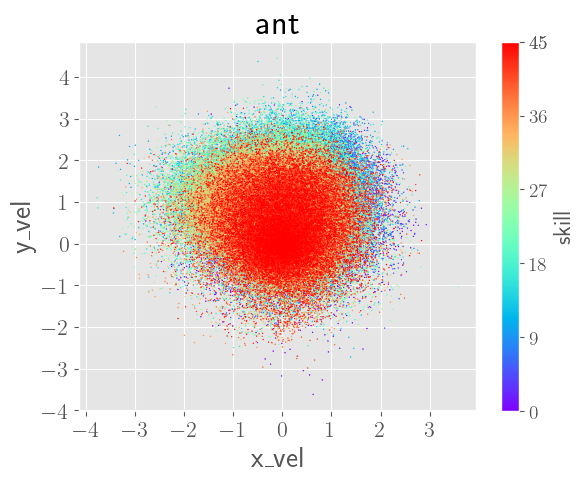}
}
\subfigure[]{\includegraphics[width=.3\textwidth]{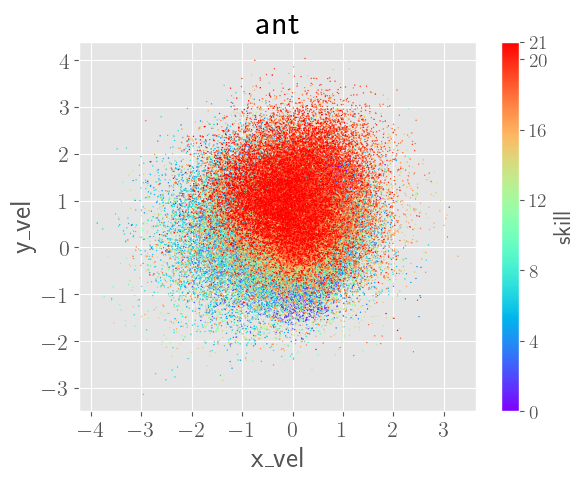}
}
\subfigure[]{\includegraphics[width=.3\textwidth]{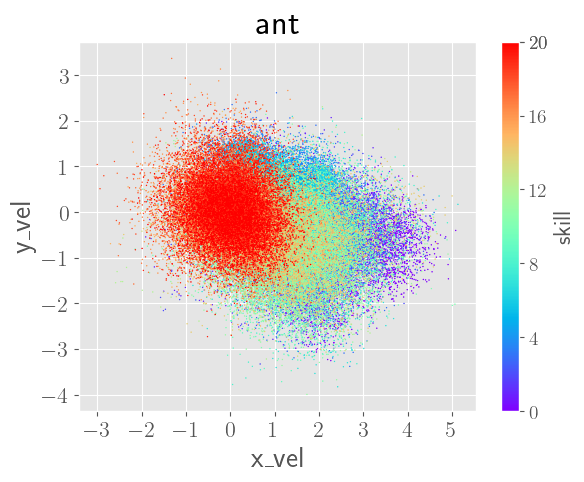}
}
\subfigure[]{\includegraphics[width=.3\textwidth]{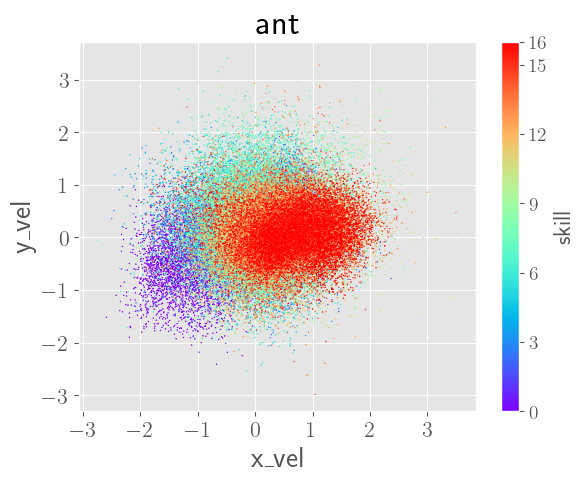}
}
\subfigure[]{\includegraphics[width=.3\textwidth]{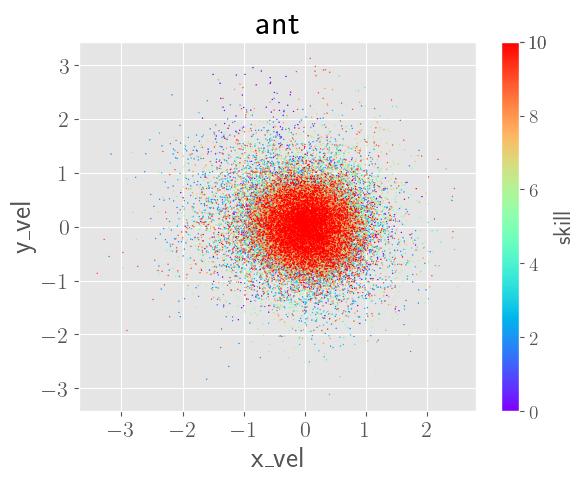}
}
\caption{Scatter-plots of the $x$- and $y$-velocity dimension of the states visited by the first $n$ skills of $6$ out of $10$ RND runs in \ant. $n$ was picked by hand in each run. The second half of the trajectory is shown. 
Colors change from early skills in purple to late skills in red.  }
\label{fig:rnd-ant-xy}
\end{center}
\vskip -0.2in
\end{figure}

\begin{figure}[ht]
\vskip 0.2in
\begin{center}
\subfigure[]{\includegraphics[width=.3\textwidth]{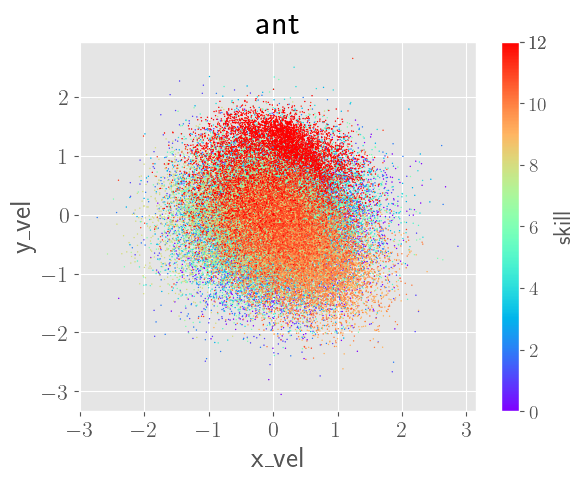}
}
\subfigure[]{\includegraphics[width=.3\textwidth]{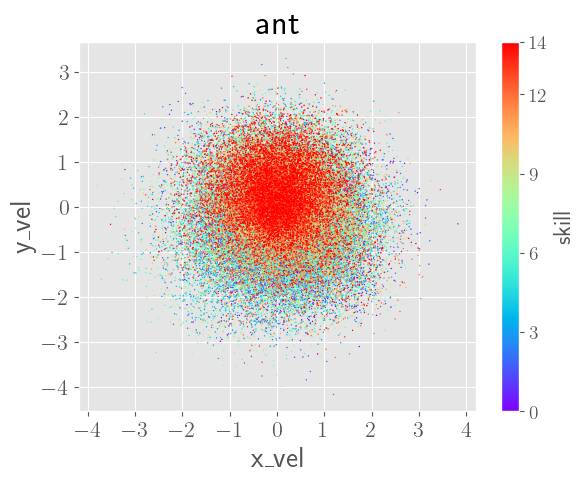}
}
\subfigure[]{\includegraphics[width=.3\textwidth]{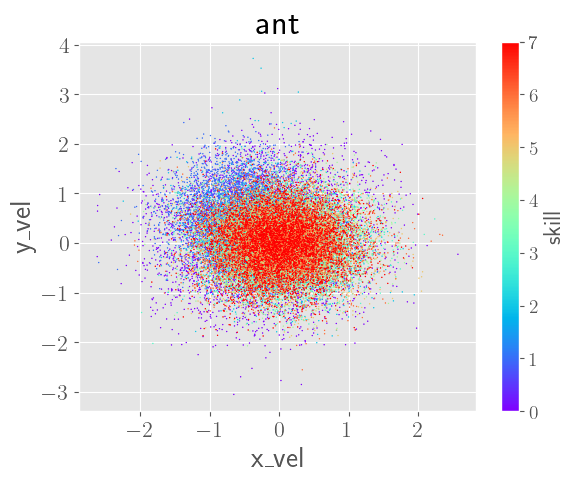}
}
\subfigure[]{\includegraphics[width=.3\textwidth]{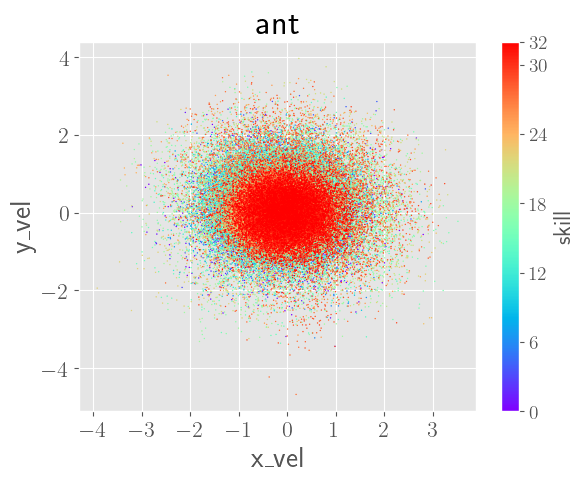}
}
\subfigure[]{\includegraphics[width=.3\textwidth]{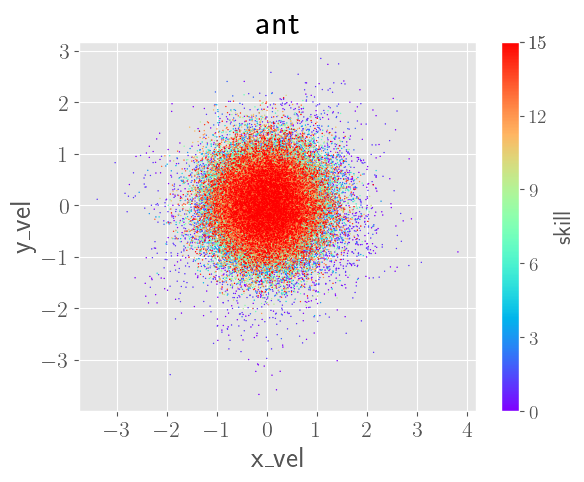}
}
\subfigure[]{\includegraphics[width=.3\textwidth]{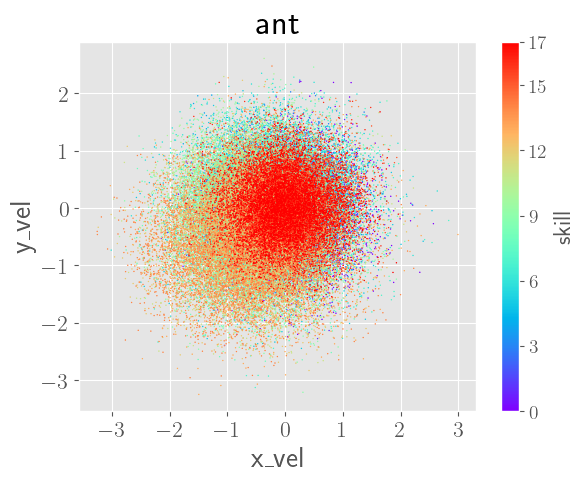}
}
\caption{Scatter-plots of the $x$- and $y$-velocity dimension of the states visited by the first $n$ skills of $6$ out of $10$ `Disagreement' runs in \ant. $n$ was picked by hand in each run. The second half of the trajectory is shown. 
Colors change from early skills in purple to late skills in red.  }
\label{fig:dis-ant-xy}
\end{center}
\vskip -0.2in
\end{figure}

\begin{figure}[ht]
\vskip 0.2in
\begin{center}
\subfigure[]{\includegraphics[width=.3\textwidth]{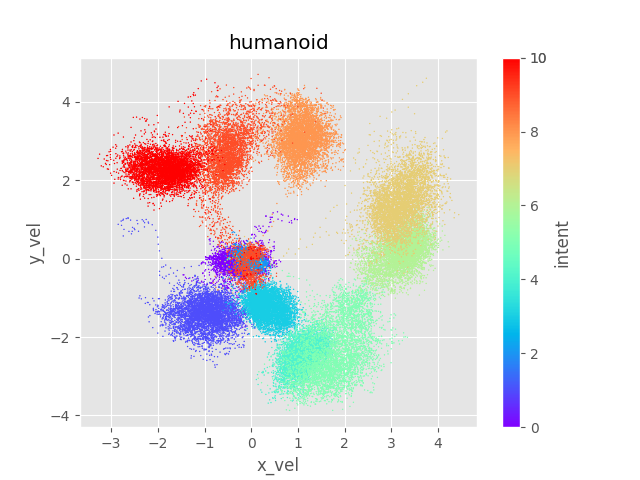}
}
\subfigure[]{\includegraphics[width=.3\textwidth]{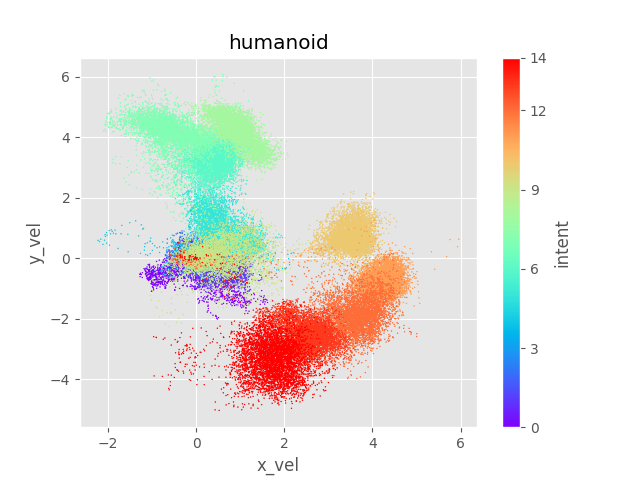}
}
\subfigure[]{\includegraphics[width=.3\textwidth]{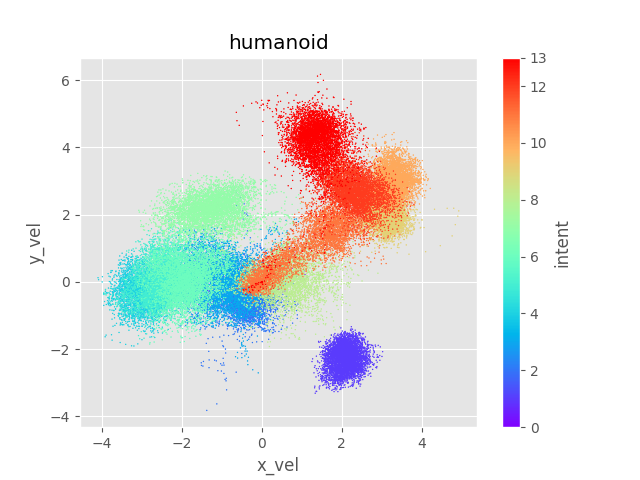}
}
\subfigure[]{\includegraphics[width=.3\textwidth]{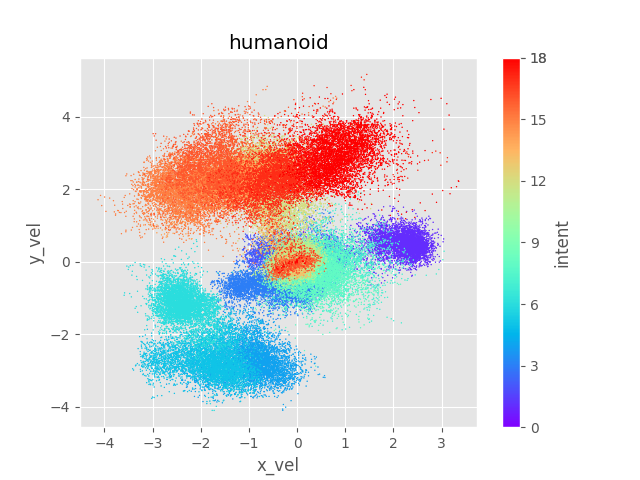}
}
\subfigure[]{\includegraphics[width=.3\textwidth]{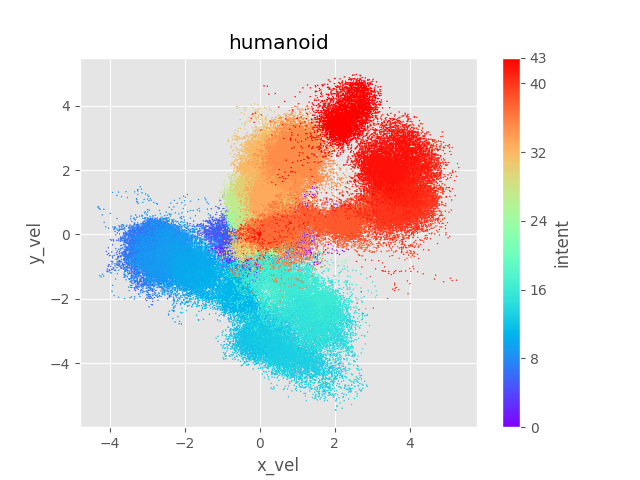}
}
\subfigure[]{\includegraphics[width=.3\textwidth]{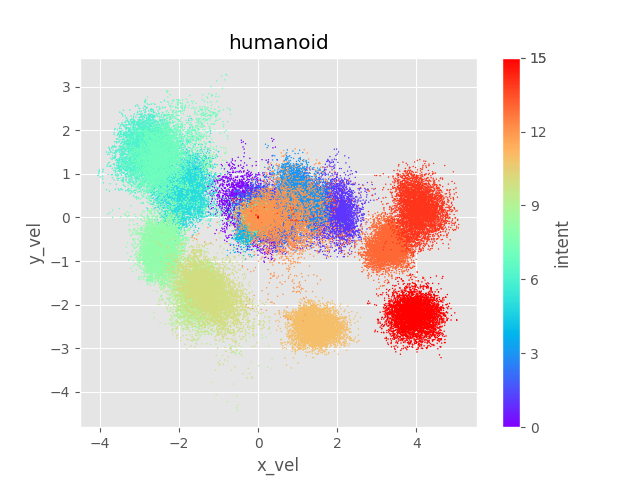}
}
\subfigure[]{\includegraphics[width=.3\textwidth]{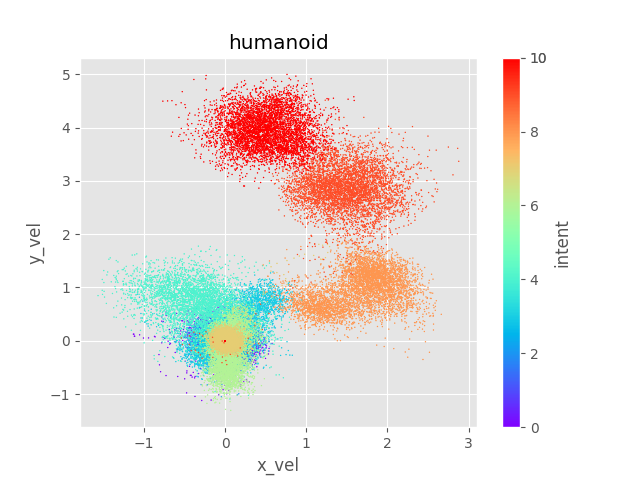}
}
\subfigure[]{\includegraphics[width=.3\textwidth]{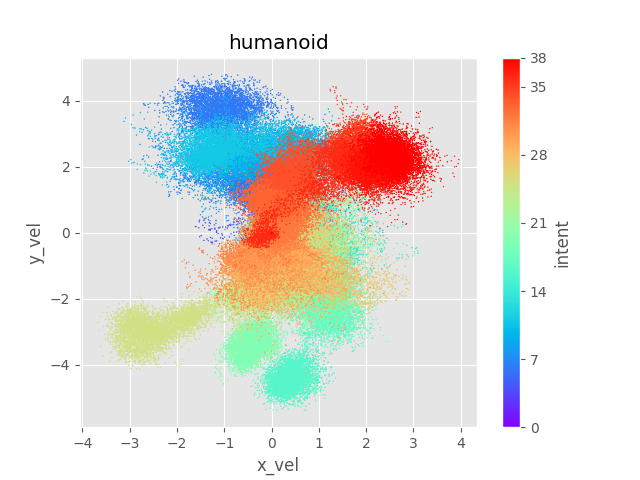}
}
\subfigure[]{\includegraphics[width=.3\textwidth]{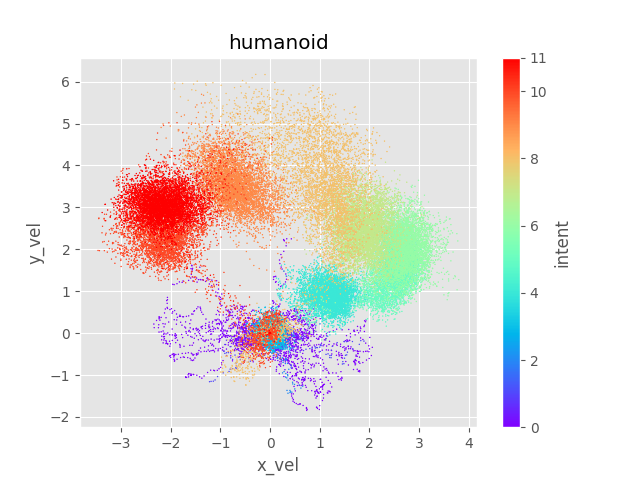}
}
\caption{Scatter-plots of the $x$- and $y$-velocity dimension of the states visited by the first $n$ skills of all additional $9$ runs in \hum. $n$ was picked by hand in each run. The second half of the trajectory is shown. 
Colors change from early skills in purple to late skills in red. }
\label{fig:app-hum-xy}
\end{center}
\vskip -0.2in
\end{figure}

\begin{figure}[ht]
\vskip 0.2in
\begin{center}
\subfigure[]{\includegraphics[width=.3\textwidth]{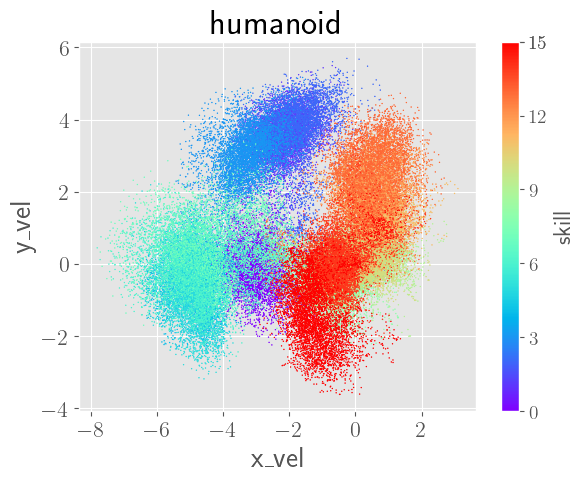}
}
\subfigure[]{\includegraphics[width=.3\textwidth]{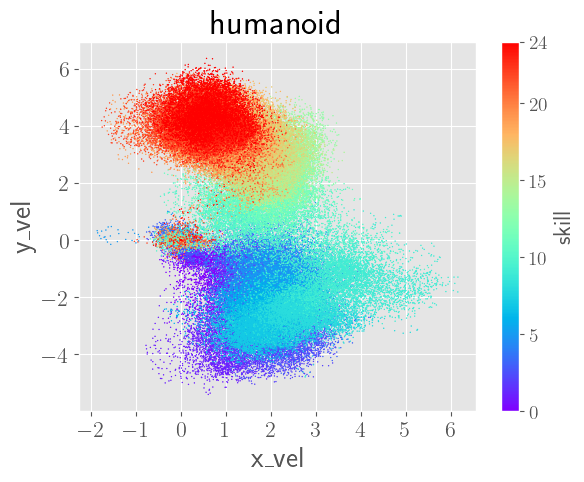}
}
\subfigure[]{\includegraphics[width=.3\textwidth]{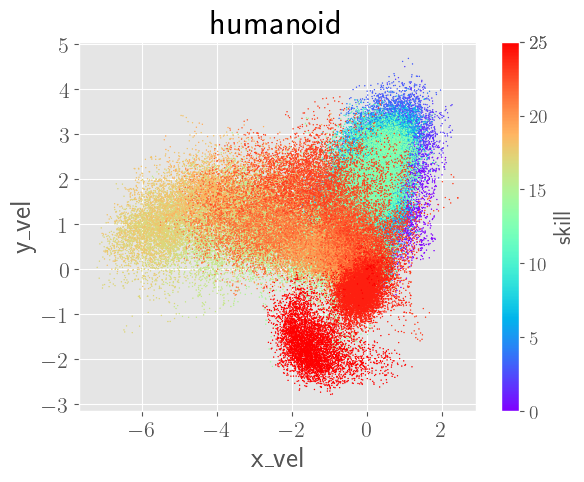}
}
\subfigure[]{\includegraphics[width=.3\textwidth]{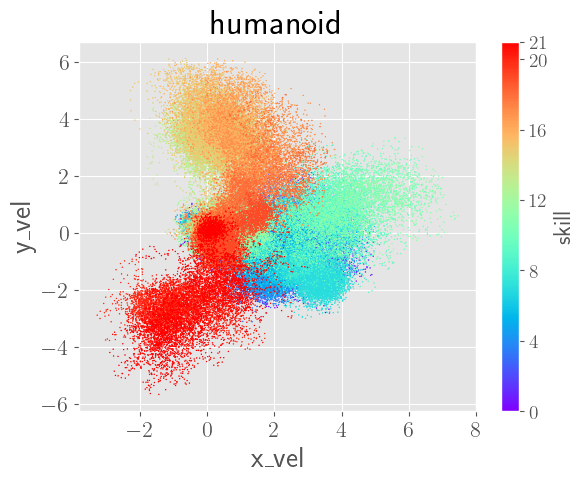}
}
\subfigure[]{\includegraphics[width=.3\textwidth]{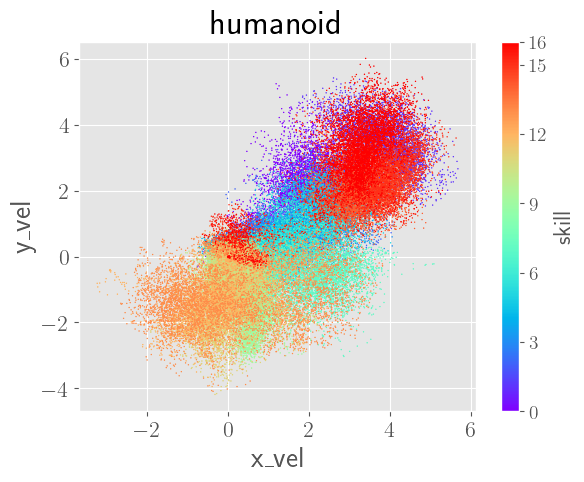}
}
\subfigure[]{\includegraphics[width=.3\textwidth]{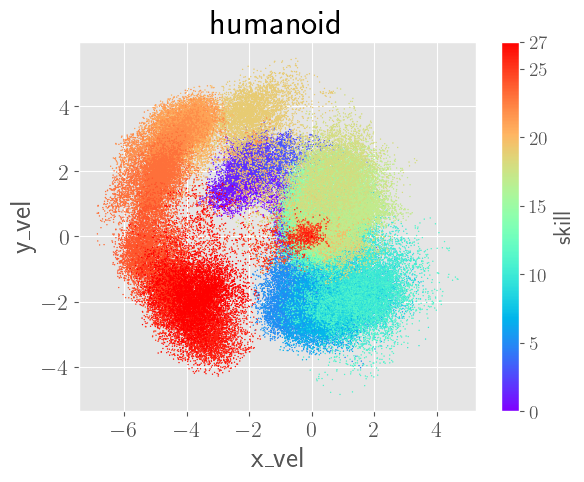}
}
\caption{Scatter-plots of the $x$- and $y$-velocity dimension of the states visited by the first $n$ skills of $6$ out of $10$ RND runs in \hum. $n$ was picked by hand in each run. The second half of the trajectory is shown. 
Colors change from early skills in purple to late skills in red. }
\label{fig:rnd-hum-xy}
\end{center}
\vskip -0.2in
\end{figure}

\begin{figure}[ht]
\vskip 0.2in
\begin{center}
\subfigure[]{\includegraphics[width=.3\textwidth]{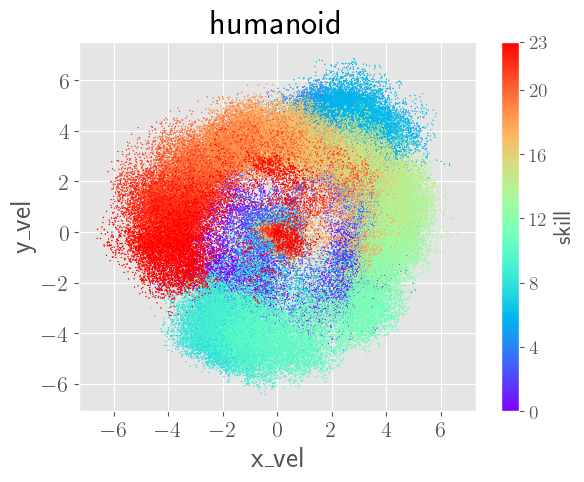}
}
\subfigure[]{\includegraphics[width=.3\textwidth]{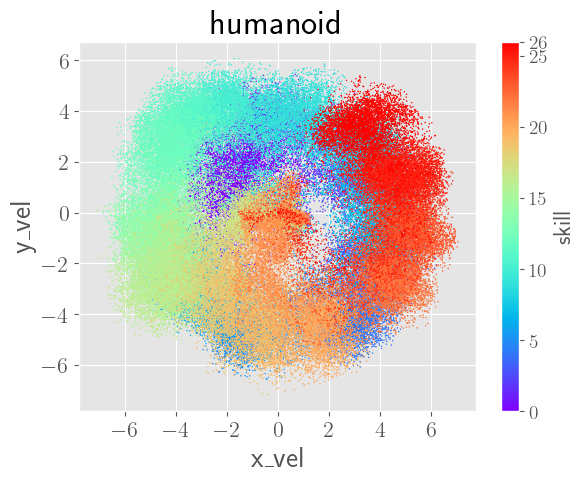}
}
\subfigure[]{\includegraphics[width=.3\textwidth]{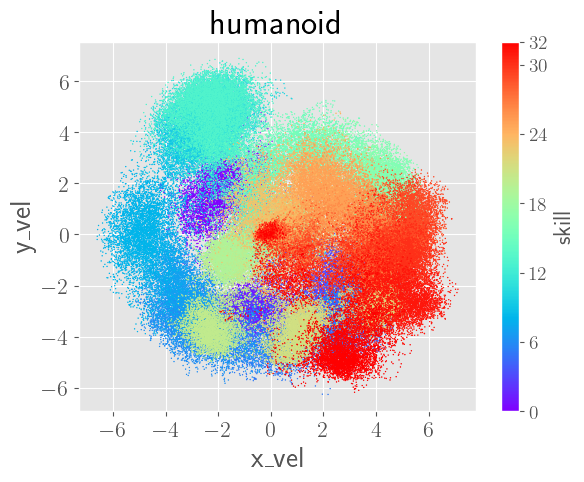}
}
\subfigure[]{\includegraphics[width=.3\textwidth]{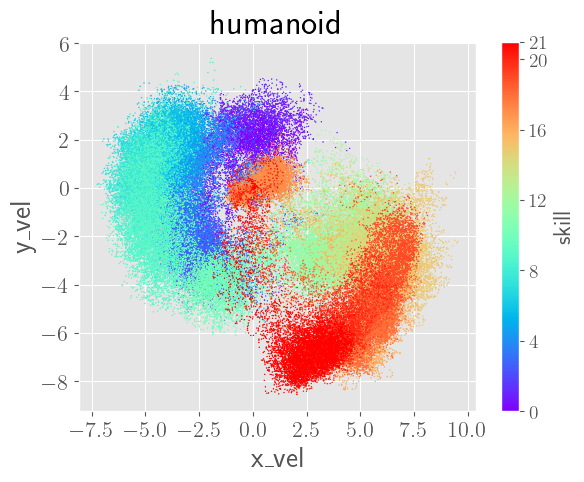}
}
\subfigure[]{\includegraphics[width=.3\textwidth]{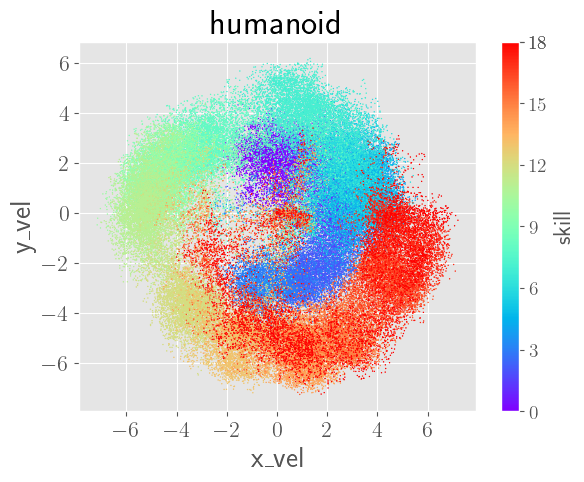}
}
\subfigure[]{\includegraphics[width=.3\textwidth]{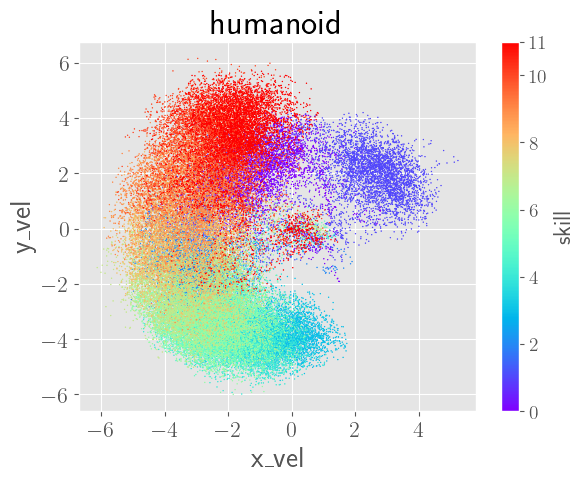}
}
\caption{Scatter-plots of the $x$- and $y$-velocity dimension of the states visited by the first $n$ skills of $6$ out of $10$ `Disagreement' runs in \hum. $n$ was picked by hand in each run. The second half of the trajectory is shown. 
Colors change from early skills in purple to late skills in red. }
\label{fig:dis-hum-xy}
\end{center}
\vskip -0.2in
\end{figure}

In Figures~\ref{fig:app-ant-xy}, \ref{fig:rnd-ant-xy} respectively \ref{fig:dis-ant-xy} we show more scatter plots for the \ant environment of our method, RND~\cite{burda2018exploration} respectively `Disagreement'~\cite{pathak2019self}.
In Figures~\ref{fig:app-hum-xy}, \ref{fig:rnd-hum-xy} respectively \ref{fig:dis-hum-xy} we show more scatter plots for the \hum environment of our method,  RND respectively `Disagreement'.

In Table~\ref{table:app-zero-shot} we report the zero-shot results for the ablations and the intrinsic curiosity baselines.
In Table~\ref{table:app-MI} we report the MI-metric results for the ablations. 
\begin{table}
\caption{Zero-shot environment reward of our algorithm, ablations and baselines.}
\label{table:app-zero-shot}
\begin{center}
  \begin{tabular}{lcc}
\toprule
Task & Method & Zero-shot  \\
 & reward\\
\midrule
Ant          & \textbf{Ours (full)} &$\mathbf{2506 \pm 511}$ \\
Ant          & Ours (policy ablation) &$1731 \pm 634$ \\
Ant          & Ours (value ablation) &$2246 \pm 794$ \\
Ant          & RND &$9 \pm 155$ \\
Ant          & Disagreement &$-170 \pm 67$ \\
\midrule
Humanoid     & Ours (full) & $9092 \pm 1063$ \\
Humanoid     & Ours (policy ablation) &$8906 \pm 616$ \\
Humanoid     & Ours (value ablation) &$7357 \pm 816$ \\
Humanoid     & RND &$7734 \pm 1916$ \\
Humanoid     & \textbf{Disagreement} & $\mathbf{10107 \pm 736}$ \\
\bottomrule
\end{tabular}
\end{center}
 \caption{Ablations for the Particle-based mutual information metric. Results are averaged over $10$ runs.}
\label{table:app-MI}
 \begin{center}
 \begin{tabular}{lccr}
\toprule
Task & Method  & MI$(s,z)$ \\
 &  & \\
\midrule
Ant          & full method &$1.33 \pm 0.11$\\
Ant          & policy ablation   &$1.09 \pm 0.16$\\
Ant          & value ablation   &$1.28 \pm 0.15$\\
\midrule
Humanoid     & full method  & $1.29 \pm 0.25$ \\
Humanoid     & policy ablation  & $0.88 \pm 0.19$ \\
Humanoid     & value ablation  & $1.01 \pm 0.09$ \\

\bottomrule
\end{tabular}
\end{center}
\end{table}

\section{Montezuma's Revenge: Adapting episode length}\label{app:montezuma-episode}

In order to save computation, we adapt the number of training steps according to several criteria.
This lets us save a lot of compute on the skills which are easy to learn. We use the following measures:

\begin{itemize}
    \item We train for $1M$ steps with guiding from the previous policy. The number of guiding steps is sampled uniformly at random between $0$ and $450$, each time the environment is reset. Then we train for another $0.65M$ steps without any guiding.
    \item If the average reward goes down after removing the guiding or is too low at any point, we restart the guiding phase for $0.65M$ steps.
    \item If the agent is reaching a terminal state in more than $10\%$ of the episodes, we continue training.
\end{itemize}

On top of this, we ignore positive samples that receive almost no reward.
All these tricks enable us to considerably reduce the training time. However, they are not a core change in our algorithm as they could all be replaced by just training all generations for a longer fixed number of steps, just as before.

\section{Montezuma's Revenge: Environment details}\label{app:montezuma-env}

\begin{figure}[ht]
\begin{center}
\subfigure[]{\includegraphics[height=4.5cm]{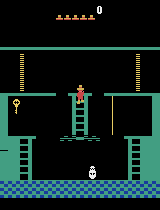}\label{fig:mz-base}
}
\subfigure[]{\includegraphics[height=4.5cm ]{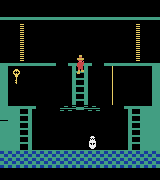}\label{fig:mz-base-cut}
}
\caption{(a) Initial state in the first room of Montezuma's Revenge. The agent controls the red/yellow character and its actions are moving up, down, right or left, and jumping. Touching the skull or jumping from a high height makes the agent lose a life. The agent must collect the key and open one of the two doors to access the next room. (b) Cropped version of the input that we use in our algorithm.}
\label{fig:montezuma-base}
\end{center}
\vskip -0.2in
\end{figure}

Figure \ref{fig:montezuma-base} shows the initial state of Montezuma's Revenge and the cropped version of it that is used for our agent.

\section{Montezuma's Revenge: Experimental details} \label{app:montezuma-exp}

In Montezuma's Revenge, we use the PPO implementation from the coax\footnote{https://github.com/coax-dev/coax} library, with 1-step temporal differences. 
If the average score per $500$ steps is below $5$, we go back to the guiding phase.
Due to the large memory requirements to store all negative samples, after the $15$-th epoch we rewrite old negative samples to add the new ones.
Table \ref{table:montezuma-hp} summarizes our hyper-parameters.

\begin{table}[ht]
\caption{Hyperparameters for Montezuma's Revenge.}
\label{table:montezuma-hp}
\vskip 0.15in
\begin{center}
\begin{small}
\begin{sc}
\begin{tabular}{lc}
\toprule
Hyper-parameter \\
\midrule
Base Entropy Regularization & $0.003$ \\
Extra Entropy Regularization & $0.03$ \\
Episode length & $500$ \\
PPO learning rate & $0.0003$ \\
PPO discount factor & $0.99$ \\
PPO epsilon & $0.2$ \\
PPO batch size & $1024$ \\
PPO replay buffer size & $4096$\\
PPO replay epochs & $4$ \\
Parallel environments & $32$\\
Positive/negative sample target value & $0.05/-0.05$ \\
\midrule
Reward network updates per skill & $1500$\\
Reward network learning rate & $0.001$ \\
Reward network training batch size & $3 \cdot 243$ \\
\bottomrule
\end{tabular}
\end{sc}
\end{small}
\end{center}
\vskip -0.15in
\end{table}

\section{Additional 2d navigation experiments}
\begin{figure}[ht]
\begin{center}
\subfigure[]{\includegraphics[height=3cm]{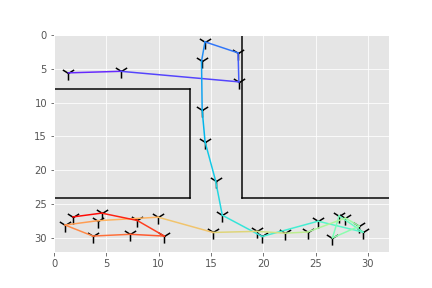}\label{fig:2d-tmaze}
}
\subfigure[]{\includegraphics[height=3cm ]{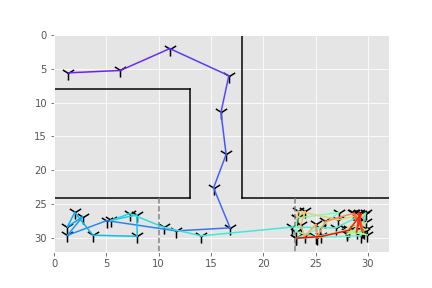}\label{fig:2d-sink}
}
\subfigure[]{\includegraphics[height=3cm ]{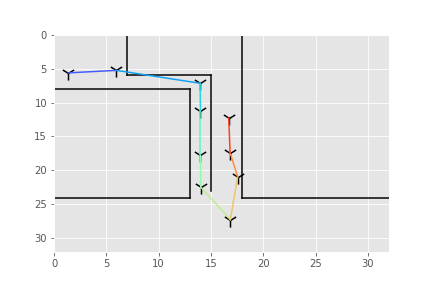}\label{fig:2d-gap}
}
\caption{The crosses correspond to the circles in Figure~\ref{fig:forward_original}. (a) A T-Maze with two possible paths. The method backtracks when it reaches the end of the path and explores the other one. (b) The dotted line can not be crossed more than once in each episode. This traps the agent and makes it impossible to backtrack. The method gets stuck in one of the two traps. (c) A narrow path that leads to an intersection.}
\label{fig:2d-t}
\end{center}
\vskip -0.2in
\end{figure}

In Figure~\ref{fig:2d-t} we show results of the 2d navigation task with three alternative environments. These environments illustrate the behavior of our algorithm in environments that require backtracking. In particular, Figure \ref{fig:2d-sink} shows a failure mode of our approach. A trap prevents the agent from backtracking and thus stops the progress of our algorithm once the search reaches that region.


\end{document}